\documentclass[pdflatex,sn-mathphys-num,iicol]{sn-jnl}% Math and Physical Sciences Numbered Reference Style
\usepackage{graphicx}%
\usepackage{multirow}%
\usepackage{amsmath,amssymb,amsfonts}%
\usepackage{amsthm}%
\usepackage{mathrsfs}%
\usepackage[title]{appendix}%
\usepackage{textcomp}%
\usepackage{manyfoot}%
\usepackage{booktabs}%
\usepackage{algorithm}%
\usepackage{algorithmicx}%
\usepackage{algpseudocode}%
\usepackage{listings}%

\usepackage[table]{xcolor}  % 提供表格着色功能
\usepackage{amsmath}
\usepackage{amssymb}
\usepackage{makecell}       % 用于改善单元格多行及垂直居中效果
\usepackage{adjustbox}
\usepackage{graphicx}       % 已加载，无需重复加载
\usepackage{multirow}       % 支持多行合并
\usepackage{booktabs}       % 更好看的表格横线
\usepackage[table]{xcolor}
\usepackage{colortbl}
\usepackage{threeparttable}
\usepackage{subcaption} % 支持 subtable
\usepackage{pifont}
\usepackage{fancyhdr}
\definecolor{deepred}{RGB}{139,0,0}
\definecolor{deepgreen}{RGB}{0,100,0}

\definecolor{reColor}{RGB}{0,0,0}
\usepackage{xspace}

\makeatletter
\DeclareRobustCommand\onedot{\futurelet\@let@token\@onedot}
\def\@onedot{\ifx\@let@token.\else.\null\fi\xspace}

\def\eg{\emph{e.g}\onedot} 
\def\ie{\emph{i.e}\onedot} 
 
\def\etc{\emph{etc}\onedot} 
 
\def\etal{\emph{et al}\onedot}
\makeatother

\theoremstyle{thmstyleone}%
\theoremstyle{thmstyletwo}%

\theoremstyle{thmstylethree}%

\begin{document}

\title[SoftHGNN: Soft Hypergraph Neural Networks for General Visual Recognition]{SoftHGNN: Soft Hypergraph Neural Networks for General Visual Recognition}


% \author[1]{\fnm{Mengqi} \sur{Lei}}\email{mengqi-lei@163.com}
% \author[2]{\fnm{Yihong} \sur{Wu}}\email{wuyihong0453@link.tyut.edu.cn}
% \author[1]{\fnm{Siqi} \sur{Li}}\email{lisiqi19971013@gmail.com}
% \author[3,4]{\fnm{Xinhu} \sur{Zheng}}\email{xinhuzheng@hkust-gz.edu.cn}
% \author[5]{\fnm{Juan} \sur{Wang}}\email{wangjuan@xjtu.edu.cn}
% \author[1]{\fnm{Yue} \sur{Gao}}\email{kevin.gaoy@gmail.com}
% \author[5]{\fnm{Shaoyi} \sur{Du}}\email{dushaoyi@xjtu.edu.cn}

\newcommand{\authorsize}{\small}

\author[1,2]{\authorsize \fnm{Mengqi} \sur{Lei}\email{mengqi-lei@163.com}}
\author[3]{\authorsize \fnm{Yihong} \sur{Wu}\email{wuyihong0453@link.tyut.edu.cn}}
% \author[1,2,\ast]{\authorsize \fnm{Siqi} \sur{Li}\email{lisiqi19971013@gmail.com}}
\author*[1,2]{\fnm{Siqi} \sur{Li}}\email{lisiqi19971013@gmail.com}
% \author[1,2]{\authorsize \fnm{Siqi} \sur{Li}\corref{cor1}\email{lisiqi19971013@gmail.com}}
\author[4,5]{\authorsize \fnm{Xinhu} \sur{Zheng}\email{xinhuzheng@hkust-gz.edu.cn}}
\author[6]{\authorsize \fnm{Juan} \sur{Wang}\email{wangjuan@xjtu.edu.cn}}
\author[6]{\authorsize \fnm{Shaoyi} \sur{Du}\email{dushaoyi@xjtu.edu.cn}}
\author[1,2]{\authorsize \fnm{Yue} \sur{Gao}\email{kevin.gaoy@gmail.com}}

\affil[\small 1]{\small \orgdiv{BNRist, THUIBCS, BLBCI, School of Software}, \orgname{Tsinghua University}, \orgaddress{\city{Beijing}, \postcode{100084}, \country{China}}}

\affil[\small 2]{\small \orgname{Yangtze Delta Region Institute of Tsinghua University, Zhejiang}, 
\orgaddress{\city{Jiaxing}, \state{Zhejiang}, \postcode{314006}, \country{China}}}

\affil[\small 3]{\small \orgdiv{Department of Mechanical Engineering}, \orgname{Taiyuan University of Technology}, \orgaddress{\city{Taiyuan}, \postcode{030024}, \country{China}}}

\affil[\small 4]{\small \orgdiv{Internet of Things Thrust, Intelligent Transportation Thrust}, \orgname{The Hong Kong University of Science and Technology (Guangzhou)}, \orgaddress{\city{Guangzhou}, \country{China}}}

\affil[\small 5]{\small \orgdiv{Department of Electronic and Computer Engineering}, \orgname{The Hong Kong University of Science and Technology}, \orgaddress{\state{Hong Kong SAR}, \country{China}}}

\affil[\small 6]{\small \orgdiv{{Department of Ultrasound, the Second Affiliated Hospital of Xi'an Jiaotong University, State Key Laboratory of Human-Machine Hybrid Augmented Intelligence, and Institute of Artificial Intelligence and Robotics}}, \orgname{Xi'an Jiaotong University}, \orgaddress{\city{Xi'an}, \country{China}}}

\fancyhf{}
\renewcommand{\headrulewidth}{0pt}   % 取消页眉横线
\fancyfoot[L]{\rule{3.7cm}{0.4pt}\\ *Corresponding Author}

\abstract{
Visual recognition relies on understanding both the semantics of image tokens and the complex interactions among them. Mainstream self-attention methods, while effective at modeling global pair-wise relations, fail to capture high-order associations inherent in real-world scenes and often suffer from redundant computation. Hypergraphs extend conventional graphs by modeling high-order interactions and offer a promising framework for addressing these limitations. However, existing hypergraph neural networks typically rely on static and hard hyperedge assignments, which lead to redundant hyperedges and overlooking the continuity of visual semantics.
% In this work, we present Soft Hypergraph Neural Networks (SoftHGNN), a plug-and-play hypergraph computation method that can be seamlessly integrated into vision encoders for general visual recognition tasks.
% that make it truly efficient and versatile in visual recognition tasks. 
\textcolor{reColor}{In this work, we present Soft Hypergraph Neural Networks (SoftHGNN), a lightweight plug-and-play hypergraph computation method for late-stage semantic reasoning in existing vision pipelines.}
Our SoftHGNN introduces the concept of soft hyperedges, where each vertex is associated with hyperedges via continuous and differentiable participation weights rather than hard binary assignments. 
These weights are produced by measuring similarities between vertex features and a small set of learnable hyperedge prototypes, yielding input-adaptive and semantically rich soft hyperedges.
% These weights are achieved by using the learnable hyperedge prototype so that the model generates semantically rich soft hyperedges.
Using soft hyperedges as the medium for message aggregation and dissemination, SoftHGNN enriches feature representations with high-order contextual associations.
% various high-order semantic relationships within visual features. 
To further enhance efficiency when scaling up the number of soft hyperedges, we incorporate a sparse hyperedge selection mechanism that activates only the top-$k$ important hyperedges, along with a load-balancing regularizer to ensure adequate and balanced hyperedge utilization. Experimental results across three tasks on five datasets demonstrate that SoftHGNN efficiently captures high-order associations in visual scenes, achieving significant performance improvements.
The source code of our framework is available at: \texttt{\url{https://github.com/Mengqi-Lei/SoftHGNN}}. 
}




\maketitle

% \begin{figure*}[t]
%     \centering
%     \includegraphics[width=0.85\linewidth]{intro.pdf}
%     \caption{The differences between the proposed soft hypergraph neural network and self-attention as well as traditional HGNN. The upper part shows the differences among the three module calculation modes. The lower left part shows the differences between paired low-order visual associations and high-order visual associations. The lower right part shows an example of visualizing the prediction results of the original version of YOLOv12 and the version with SoftHGNN introduced.}
%     \label{fig:intro}
%     \vspace{-10pt}
% \end{figure*}

\section{Introduction}

In recent years, significant advancements in visual recognition have primarily been driven by deep neural network-based models, especially Convolutional Neural Networks (CNNs) and Transformer models \cite{ijcv_review,unsuper_vision_survey,vit_survey}. CNNs excel at capturing local spatial patterns, whereas Transformer models, built upon self-attention mechanisms, effectively model global relationships among image tokens. Current mainstream paradigms, such as Vision Transformer (ViT) \cite{vit} and its variants \cite{pvt,swin,LRT_tip, eatformer_ijcv, gridformer_ijcv}, typically process images into sequences of visual tokens, emphasizing the modeling of pair-wise relations between these tokens.

Despite achieving remarkable success, vision models based on self-attention mechanisms \cite{attn,vit} still face significant challenges. Firstly, self-attention mechanisms inherently construct a fully connected semantic graph that primarily focuses on pair-wise relationships among tokens, making it difficult to effectively represent complex high-order semantic associations commonly present in real-world visual scenes. 
% For instance, the simple scenario of ``two people playing badminton" involves interactions among multiple entities such as individuals, badminton rackets, and the shuttlecock. 
% For instance, as shown in Figure \ref{fig:intro}, 
For instance, the simple scene of ``a person holding a tennis racket and hitting the ball" involves the interaction among multiple entities, such as the connection among the person, the tennis racket, and the sports ball. 
Such high-order associations cannot be adequately captured by merely modeling pair-wise relations. Secondly, Transformers typically employ dense global self-attention, leading to substantial redundant computations, high computational costs, and it has been demonstrated \cite{attn_collapse1,attn_collapse2,attn_collapse3} that attention collapse and dispersion may occur in deep neural networks.

To alleviate these issues, hypergraph, as an extension of traditional graph structure, has received considerable attention due to its intrinsic capability to explicitly model high-order associations \cite{hg_book, hg_survey}. Hypergraph Neural Networks (HGNNs) explicitly represent relationships among multiple vertices via hyperedges, making them promising for capturing complex semantic interactions in visual data \cite{hgnn, hg_survey4}. Recent studies, such as Hyper-YOLO \cite{hyperyolo}, have initially validated the effectiveness of HGNNs in various vision tasks. However, existing HGNN methods, originally designed for network-type data, encounter critical issues when applied to visual tasks:
\textbf{1) Redundant hyperedges}. Traditional hypergraph constructions typically employ methods like $k$-Nearest Neighbors or $\epsilon$-ball criteria, which generate one hyperedge centered around each vertex. In visual tasks, given the large number of vertices (typically $B \times H \times W$, where $B$ is the batch size), this approach often generates an excessively large number of hyperedges, resulting in a highly-overlapping hyperedge set, where many hyperedges are near-duplicates and contribute little additional information while substantially increasing computation for the target task.
\textbf{2) Limitations of hard hyperedges}. Existing HGNNs typically rely on hard vertex-to-hyperedge associations, meaning a vertex is either completely included in a hyperedge or entirely excluded from it. This hard binary assignment overlooks the inherent continuity and ambiguity of visual semantics. For example, in partially occluded scenarios, regions of occluded objects should ideally participate in relevant hyperedges with lower weights, rather than being completely excluded. Moreover, hard hyperedge designs often cause vertex redundancy or incomplete coverage, thereby introducing noisy structures and negatively impacting model performance \cite{hg_quality,hg_quality2, hg_quality3}.

% To overcome these bottlenecks, we propose the Soft Hypergraph Neural Network (SoftHGNN), a plug-and-play hypergraph computation method that can be seamlessly integrated into vision encoders for general visual recognition tasks.
\textcolor{reColor}{To overcome these bottlenecks, we propose Soft Hypergraph Neural Networks (SoftHGNN), a lightweight plug-and-play hypergraph computation method for late-stage semantic reasoning in existing vision pipelines.}
% Figure \ref{fig:intro} shows the differences in the computing patterns of self-attention, traditional HGNN and the proposed SoftHGNN. 
The core innovation is transitioning from static and hard hyperedge assignments to a dynamic and feature-driven approach with a differentiable soft vertex-participation mechanism.  Specifically, different from traditional hyperedge construction methods based on geometric distances or feature similarities, SoftHGNN employs a set of learnable hyperedge prototype vectors. By measuring the similarity between visual tokens and these prototype vectors, the model dynamically generates semantically rich ``soft hyperedges". Each hyperedge softly connects all vertices with learnable participation weights, thus adaptively modeling abstract high-order semantic relationships within visual features. Furthermore, SoftHGNN only needs to maintain a constant-scale set of soft hyperedges, significantly enhancing computational efficiency and effectiveness compared to traditional HGNNs. 
% Moreover, as a plug-and-play module that can be easily integrated into the model of any visual recognition task to fill the gap in the ability of existing methods to model high-order visual semantic associations, thereby bringing about significant performance improvements.

In certain complex scenarios, a larger number of soft hyperedges may be required to comprehensively capture visual semantic associations. However, only a subset of these hyperedges is typically critical to the recognition task. Therefore, to further enhance the modeling capability of SoftHGNN while maintaining computational efficiency, we introduce a sparse hyperedge selection mechanism. Specifically, we pre-define a relatively large set of soft hyperedges, but select only the most important $k$ hyperedges for message passing. Moreover, we propose a load-balancing regularizer to prevent selection imbalances, \eg, hyperedge inactivity or excessive activation, ensuring adequate and balanced hyperedge utilization.

\textcolor{reColor}{
Our SoftHGNN can be easily integrated into models as a late-stage semantic reasoning block for a wide range of visual recognition tasks to fill the gap in their ability to model high-order visual semantic associations, thereby bringing about significant performance improvements.}
To validate the generalization and effectiveness of our proposed method, extensive experiments are conducted on five mainstream datasets, namely ImageNet-100 \cite{imgnet}, ImageNet-1k \cite{imgnet}, ShanghaiTech Part-A \cite{shanghaitech}, ShanghaiTech Part-B \cite{shanghaitech}, and MS COCO \cite{mscoco}.
These experiments cover three representative visual recognition tasks: image classification, crowd counting, and object detection. Comprehensive results demonstrate that the proposed SoftHGNN can accurately and efficiently capture high-order semantic relationships in visual scenes, achieving significant performance improvements.

In summary, our main contributions are as follows:
\begin{itemize}
    \item We propose the SoftHGNN, a plug-and-play hypergraph computation method to enhance vision models for general visual recognition. By introducing a feature-driven soft hyperedge mechanism, our model adaptively captures abstract high-order semantic relationships within visual features.
    \item We design a sparse hyperedge selection strategy that expands SoftHGNN's capacity by scaling up the number of hyperedges while maintaining high computational efficiency. Additionally, we incorporate a load-balancing regularizer to avoid hyperedge selection imbalance.
    \item We conduct extensive experiments on five datasets across three mainstream visual recognition tasks. Experimental results demonstrate that our method accurately and efficiently captures high-order semantic associations in visual scenes, resulting in significant performance improvements.
\end{itemize}

\section{Related Work}

\subsection{Hypergraph Computation}

The hypergraph is a kind of structure that naturally represents groups and high-order relationships within a system. Formally, it is an extension of a conventional graph, in which a single hyperedge can connect an arbitrary number of vertices. This structure is particularly well-suited for describing group interactions \cite{hg_book}. For example, in a co-authorship network, a paper can be considered a hyperedge that connects all the authors involved. Similar high-order relationships are widely present in fields including biology and neuroscience \cite{hg_survey, hg_survey2}.

Hypergraph Neural Networks (HGNNs) incorporate neural networks into hypergraph representation learning. HGNNs leverage hypergraph convolution, a two-step message-passing scheme that first aggregates vertex features into hyperedges and then redistributes them back to vertices, to capture high-order correlations \cite{hg_book,hg_survey4,hyper3dg}. Compared to traditional Graph Neural Networks (GNNs), this high-order information aggregation mechanism significantly enhances the representational power of HGNNs \cite{hg_book,hg_survey3}. Feng \etal~\cite{hgnn} proposed a spectral-domain hypergraph convolution method that demonstrated outstanding performance on high-order relational data. Building on this, Gao \etal~\cite{hgnnp} introduced a more flexible approach for modeling higher-order relationships along with a generalized spatial hypergraph convolution framework, extending both the adaptability of hypergraph modeling and the universality of hypergraph convolution operations.

Recently, several works have introduced attention or Transformer-style mechanisms to parameterize vertex-hyperedge interactions. 
Jiang \etal \cite{dhgnn} proposed the Dynamic Hypergraph Neural Network, which dynamically updates the hypergraph structure across layers, aiming to better reflect hidden relations not captured by an initially constructed structure.
Kim \etal \cite{han} introduced the Hypergraph Attention Network (HAN). HAN constructs hypergraphs from symbolic structures and uses co-attention between hyperedges to integrate multimodal representations.
Li \etal \cite{hgtn} designed the Hypergraph Transformer Neural Network (HGTN), which explores Transformer-based attention on heterogeneous information networks to capture higher-order semantics and facilitate communication between vertex and hyperedge types.

Hypergraph learning has shown significant advantages in modeling high-order relationships across various domains such as social networks \cite{social,social2}, knowledge graphs \cite{kg}, recommendation systems \cite{recom,recom2}, action recognition \cite{ActRec_tip,ActRec_tip2}, and bioinformatics \cite{bio,bio2}.

% It is important to note, however, that these tasks are inherently based on non-Euclidean data with graph or hypergraph structures. Hypergraphs have not yet been widely applied to data in explicit Euclidean spaces, such as visual data like images.

% 
% \subsection{Attention-based Hypergraph Models}
% Beyond convolutional hypergraph message passing, several works introduce attention or Transformer-style mechanisms to parameterize vertex-hyperedge interactions.
% HAN \cite{han_multimodal} constructs hypergraphs from symbolic structures and uses co-attention between hyperedges to integrate multimodal representations.
% HGTN \cite{hgtn} explores Transformer-based attention on hypergraph/heterogeneous information networks to capture higher-order semantics and facilitate communication between node and hyperedge types.
% DHGNN \cite{dhgnn} dynamically updates the hypergraph structure across layers (via a dynamic construction module) and then performs hypergraph convolution, aiming to better reflect hidden relations not captured by an initially constructed structure.
% \emph{Relation to our work:} SoftHGNN also produces attention-like continuous weights for vertex participation in hyperedges.
% \emph{Key differences:} our method targets visual token representations inside standard vision encoders and constructs \emph{soft hyperedges} using a small set of learnable prototypes, yielding a fully differentiable participation matrix via column-wise Softmax, rather than relying on task-specific symbolic graphs (HAN) or general graph/hypergraph priors and dynamic rebuilding (DHGNN/HGTN).
% 

\subsection{Hypergraph Neural Networks in Visual Recognition}
In recent years, research has gradually begun to explore the incorporation of hypergraphs and hypergraph neural networks into visual recognition tasks \cite{ActRec_tip,ActRec_tip2,3d_retr_tip}. Feng \etal~proposed Hyper-YOLO \cite{hyperyolo}, which integrates a hypergraph computation module into the neck component of YOLO \cite{yolo} to achieve cross-level and cross-spatial hypergraph semantic aggregation. Han \etal~introduced Vision HGNN \cite{vihgnn}, exploring the replacement of Transformer modules with hypergraph convolution modules. Wang \etal~proposed HGFormer \cite{hgformer}, which establishes many-to-many topological connections at the patch level and utilizes a hypergraph attention mechanism to learn global visual relationships. Chen \etal~proposed HI2R \cite{intra_inter}, which simultaneously extracts intra-category structural information and inter-category relational information through hypergraphs for fine-grained recognition.

\textcolor{reColor}{
Most recently, Fixelle \etal proposed Hypergraph Vision Transformers (HgVT) \cite{hgvt}, a dedicated vision backbone that unifies hypergraph modeling with Transformer-style attention. HgVT dynamically computes a soft vertex-hyperedge adjacency matrix using a sharpened sigmoid. It then derives a hard adjacency mask for sparse vertex self-attention, while retaining the soft adjacency to modulate the vertex-to-hyperedge and hyperedge-to-vertex communication pathways. HgVT further introduces population regularization to shape hyperedge membership density and adopts an expert-based pooling strategy for prediction. Notably, its vertex-hyperedge communication remains differentiable through the soft adjacency matrix.}

\textcolor{reColor}{
Although hypergraph-based visual recognition has made progress, many existing approaches still face limitations in representing visual semantics. Traditional methods \cite{vihgnn,hyperyolo} relying on $k$-NN or $\epsilon$-ball suffer from inherent hyperedge redundancy and quadratic computational costs. Recent attention-based methods such as HAN \cite{han} and HGTN \cite{hgtn} have introduced more flexible frameworks, but the process of constructing the hypergraph in these methods remains non-learnable. HgVT \cite{hgvt} further advances the area by combining sparse vertex self-attention with soft vertex-hyperedge communication. In this paper, we explore a different design point: SoftHGNN is a lightweight add-on block that performs vertex $\rightarrow$ hyperedge $\rightarrow$ vertex message passing directly over a continuous participation matrix, without introducing a separate thresholded vertex-to-vertex self-attention branch. In addition, we scale up the hyperedge capacity efficiently through sparse hyperedge selection with load-balancing regularization.}

\section{Preliminaries on Traditional Hypergraph}
\subsection{Hypergraph}

A hypergraph is an extension of a graph that explicitly models high-order relationships within data. Formally, a hypergraph is represented as:
$
\mathcal{G} = (\mathcal{V}, \mathcal{E}),
$
where $\mathcal{V}$ denotes the set of vertices, and $\mathcal{E}$ denotes the set of hyperedges. Unlike traditional graphs, each hyperedge in a hypergraph can simultaneously connect an arbitrary number of vertices. As a result, hypergraphs naturally capture complex high-order associations among multiple vertices. The structure of a hypergraph can be described using an incidence matrix ${H} \in \mathbb{R}^{|\mathcal{V}|\times|\mathcal{E}|}$, where $H_{v,e}=1$ if $v \in e$ and $H_{v,e}=0$ otherwise.
% $
% H_{v,e}=
% \begin{cases}
% 1,&\text{if vertex } v \in e,\\
% 0,&\text{otherwise}.
% \end{cases}
% $

\subsection{Hypergraph Neural Networks}

The key of Hypergraph Neural Networks (HGNNs) is to capture complex semantic relationships by facilitating information interaction between vertices and hyperedges, thereby obtaining richer feature representations. An HGNN consists of multiple layers of hypergraph convolution. A hypergraph convolution operation includes two information aggregation steps: vertex-to-hyperedge aggregation and hyperedge-to-vertex aggregation. 
Given a vertex feature matrix $X^t \in \mathbb{R}^{|\mathcal{V}|\times d}$, a typical hypergraph convolution operation is formulated as:
\begin{equation}
X^{t+1} = \sigma(D_v^{-1} H W D_e^{-1} H^\top X^{t}\Theta^{t}),
\end{equation}
where $X^t$ and $X^{t+1}$ represent the vertex feature matrices at layer $t$ and $t+1$, respectively.
$H$ is the incidence matrix of the hypergraph.
$W \in \mathbb{R}^{|\mathcal{E}|\times|\mathcal{E}|}$ is the hyperedge weight matrix, which is typically set as an identity matrix.
$\Theta^{t}$ is the learnable weight matrix, responsible for feature transformation in the hypergraph convolution.
$\sigma(\cdot)$ is a nonlinear activation function.
$D_v$ and $D_e$ are the vertex degree matrix and hyperedge degree matrix, respectively, used for normalizing the information propagation. They are defined as: 
$
    (D_v)_{i,i} = \sum_{e \in \mathcal{E}} H_{i,e}, \quad (D_e)_{j,j} = \sum_{v \in \mathcal{V}} H_{v,j}.
$

Despite the theoretical promise of HGNNs in capturing high-order semantic associations, several inherent limitations arise when applying them to visual recognition tasks. First, traditional hypergraph constructions typically rely on methods such as \(k\)-Nearest Neighbors or \(\epsilon\)-ball criteria, which generate one hyperedge per vertex. In visual applications, where the number of vertices is extremely large, this approach leads to an explosion of hyperedge numbers. 
This redundancy not only introduces substantial computational overhead but also results in many hyperedges that capture overlapping or less task-relevant relationships, thereby diluting the model’s focus on truly meaningful associations.
% The redundancy not only introduces substantial computational overhead but also results in many hyperedges that capture overlapping or irrelevant relationships, thereby diluting the model's focus on truly meaningful associations.
Second, the incidence matrix \(H\) in the traditional hypergraph enforces a hard binary assignment: a vertex is either fully included in a hyperedge or not at all. Such a rigid formulation fails to capture the inherent continuity and ambiguity present in visual data, especially in scenarios with partial occlusions or overlapping semantic regions, thus limiting the expressiveness and robustness of the model.

\subsection{Complexity Analysis}

In the process of modeling hypergraphs and HGNNs, computational complexity is a crucial concern. Especially when dealing with large-scale datasets, both the traditional methods of hypergraph construction and the operations within HGNNs can impose significant computational and time costs. In the following, we analyze the complexity of classical hypergraph construction methods (based on \(k\)-Nearest Neighbors and \(\epsilon\)-ball) as well as that of HGNNs.

\subsubsection{Complexity of Hypergraph Construction}

Common hypergraph construction methods mainly include the \(k\)-Nearest Neighbors (\(k\)-NN) approach and the \(\epsilon\)-ball approach. Both methods require searching for neighborhood relationships among vertices, leading to high computational overhead.

\textbf{\(k\)-NN}.
For a dataset with \(N\) vertices, the \(k\)-NN method requires computing the distances between all pairs of vertices, resulting in a complexity of \(\mathcal{O}(N^2\times D)\), where $D$ is the dimension of the features. 
% Although acceleration techniques like kd-trees \cite{kd_tree} can be applied in low-dimensional spaces, these methods tend to offer limited improvements in high-dimensional settings due to the ``curse of dimensionality". Therefore, constructing a hypergraph based on \(k\)-NN is computationally expensive.

\textbf{\(\epsilon\)-Ball}.  
The \(\epsilon\)-ball approach involves checking each pair of vertices to determine whether their distance is below a predefined threshold \(\epsilon\), so it also has a complexity of \(\mathcal{O}(N^2 \times D)\). 
% In high-dimensional spaces, where the distribution of distances is often more uniform, selecting an appropriate \(\epsilon\) becomes more challenging, which may result in a significant increase in the number of hyperedges and further exacerbate the computational burden.

Thus, whether using \(k\)-NN or \(\epsilon\)-ball, the primary bottleneck lies in pairwise distance computations among vertices, leading to extreme computational cost for large-scale data.

% \begin{figure*}[t]
%     \centering
%     \includegraphics[width=1\linewidth]{Framework.pdf}
%     \caption{Overview of the Soft Hypergraph Neural Network (SoftHGNN) pipeline. (1) Soft Hyperedge Generation. A set of learnable prototype vectors is dynamically offset by a context‑aware network to form sample‑specific soft hyperedges. Then each vertex computes participation weights to build the continuous participation matrix $A$. (2) Message Passing on Soft Hypergraph. High‑order interactions are captured via a two‑stage aggregation from vertices to hyperedges ($\mathcal{V}\!\to\!\mathcal{E}$) and from hyperedges to vertices ($\mathcal{E}\!\to\!\mathcal{V}$). (3) Sparse Hyperedge Selection (optional). The sparse hyperedge selection strategy is introduced to expand the number of hyperedges while avoiding a significant increase in computational complexity.}
%     \label{fig:framework}
%     % \vspace{-0.4cm}
% \end{figure*}

\begin{table*}[t]
\centering
\small
\setlength{\tabcolsep}{13pt}
\renewcommand{\arraystretch}{1.15}
\caption{Important symbols used in the Proposed Method section.}
\label{tab:symbol}
\begin{tabular}{ll}
\toprule
\textbf{Symbol}& \textbf{Description}\\
\midrule
$G$ & soft hypergraph \\
$V=\{v_i\}_{i=1}^{N}$ & vertex set \\
$E=\{e_m\}_{m=1}^{M}$ & soft hyperedge set (each $e_m$ conceptually involves all vertices) \\
$N$ & number of vertices in a sample \\
$M$ & preset number of soft hyperedges \\
$D$ & vertex feature dimension \\
$X\in\mathbb{R}^{N\times D}$& vertex feature matrix; row $i$ is $x_i$ \\
$x_i\in \mathbb{R}^{D}$& feature of vertex $v_i$ \\
$A\in [0,1]^{N\times M}$& participation matrix; $A_{i,m}$ is participation weight of $v_i$ in $e_m$ \\
$H \in \{0,1\}^{N\times M}$& binary incidence matrix in traditional hypergraph \\
\midrule
$P_0\in \mathbb{R}^{M\times D}$& globally shared learnable hyperedge prototype parameters \\
$\Delta P\in \mathbb{R}^{M\times D}$& sample-specific prototype offset\\
$P\in \mathbb{R}^{M\times D}$& dynamic hyperedge prototype (sample-specific)\\
$S\in \mathbb{R}^{N\times M}$& unnormalized prototype-vertex feature similarity score.\\
\midrule
$\mathcal{E}_{\text{fixed}},\mathcal{E}_{\text{dyn}}$ & fixed and dynamic soft hyperedge subsets (Eq.~(18))\\
$M_{\text{fixed}},M_{\text{dyn}}$ & sizes of $E_{\text{fixed}}$ and $E_{\text{dyn}}$; $M=M_{\text{fixed}}+M_{\text{dyn}}$ \\
$S_{\text{dyn}}\in \mathbb{R}^{N\times M_{\text{dyn}}}$& columns of $S$ corresponding to dynamic soft hyperedges \\
$g_j$ & activation score of dynamic hyperedge $j$ (Eq.~(19))\\
$k$ & number of selected dynamic hyperedges (top-$k$) \\
$\mathcal{E}_{\text{sel}}$ & selected dynamic hyperedge subset, $|\mathcal{E}_{\text{sel}}|=k$ \\
$S'\in \mathbb{R}^{N\times (M_{\text{fixed}}+k)}$& updated unnormalized assignments after selection \\
$A_{\text{total}}\in [0,1]^{N\times (M_{\text{fixed}}+k)}$& final participation matrix after selection and Softmax (Eq.~(20))\\
$T$ & number of forward passes used to compute activation probabilities \\
$\mathcal{M}\in \{0,1\}^{T\times M_{\text{dyn}}}$& binary selection mask over dynamic hyperedges across passes (Eq.~(21))\\
$p_j\in [0,1]$& activation probability of dynamic hyperedge $j$ (Eq.~(22))\\
$p_{\text{target}}\in [0,1]$& target activation probability.\\
$\mathcal{L}_{\text{LB}}$ & load-balancing regularization loss (Eq.~(23))\\
\bottomrule
\end{tabular}
\end{table*}

\subsubsection{Complexity of Hypergraph Neural Networks}

In the operation of HGNNs, the main computational costs come from the following matrix multiplications:

\textbf{\(H^\top X^t\):}  
    Multiplying the transpose of the incidence matrix \(H^\top\) (of size \(M \times N\)) with the vertex feature matrix \(X^t\) (of size \(N \times D\)) incurs a complexity of \(\mathcal{O}(N \times M \times D)\).
    
\textbf{\(H W D_e^{-1}\):}  
    If the hyperedge weight matrix \(W\) is a diagonal or a sparse matrix, the complexity mainly depends on the sparsity of \(H\); however, in the worst case, this multiplication may also reach \(\mathcal{O}(N \times M)\).
    
\textbf{Overall Computation:}  
    Consequently, the overall complexity of the hypergraph convolution operation is approximately \(\mathcal{O}(N \times M \times D)\).

It is important to note that in practice, because hypergraphs are usually constructed by approaches such as \(k\)-NN or \(\epsilon\)-ball, the number of hyperedges \(M\) is typically the same as the number of vertices \(N\). Therefore, the overall computational complexity of HGNNs can often be regarded as \(\mathcal{O}(N^2)\).

In summary, both traditional hypergraph construction methods and hypergraph neural networks incur significant computational overhead, and the complexity reaches or even far exceeds \(\mathcal{O}(N^2)\). This high computational cost, especially when dealing with large-scale and high-dimensional data, highlights a major efficiency bottleneck. Consequently, the intensive computational demands hinder the widespread application of hypergraph-based approaches in fields like visual recognition, where efficiency is critical.

\begin{figure*}[t]
    \centering
    \includegraphics[width=1\linewidth]{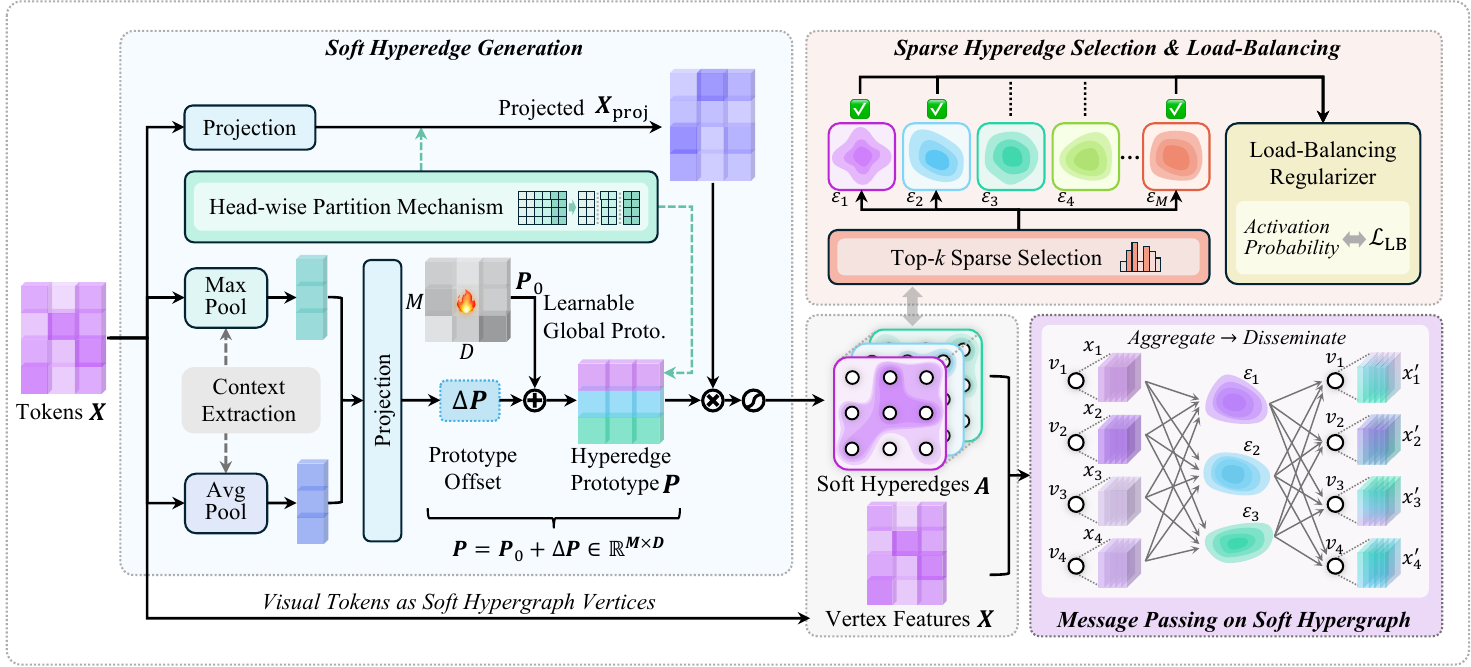}
    \caption{Overview of the Soft Hypergraph Neural Network (SoftHGNN) pipeline. (1) Soft Hyperedge Generation. A set of learnable prototype vectors is dynamically offset by a context‑aware network to form sample‑specific soft hyperedges. Then each vertex computes participation weights to build the continuous participation matrix $A$. (2) Message Passing on Soft Hypergraph. High‑order interactions are captured via a two‑stage aggregation from vertices to hyperedges ($\mathcal{V}\!\to\!\mathcal{E}$) and from hyperedges to vertices ($\mathcal{E}\!\to\!\mathcal{V}$). (3) Sparse Hyperedge Selection (optional). The sparse hyperedge selection strategy is introduced to expand the number of hyperedges while avoiding a significant increase in computational complexity.}
    \label{fig:framework}
    % \vspace{-0.4cm}
\end{figure*}

\section{Proposed Method}

In this section, we present the proposed SoftHGNN in six parts. Section \ref{sec-a} gives the definition of a Soft hypergraph. Section \ref{sec-b} provides an overall introduction to the SoftHGNN workflow. Section \ref{sec-c} and Section \ref{sec-d} describe the construction of the soft hypergraph and the message passing on the soft hypergraph, respectively. Section \ref{sec-e} introduces sparse soft hyperedges and the load-balancing regularization algorithm. Section \ref{sec-f} discusses the computational complexity of SoftHGNN. Table \ref{tab:symbol} lists the main symbols and their descriptions for easy reference.

\subsection{Definition of Soft Hypergraph}
\label{sec-a}

The design of soft hypergraphs is fundamentally different from that of traditional hypergraphs in the way they model relationships. 
% In traditional hypergraphs, hyperedge construction typically relies on predefined rules (\eg, thresholding based on geometric distances or similarity measures), and the membership is determined in a hard binary manner, with each hyperedge containing only a few vertices. This explicit and rigid definition fails to capture the pervasive ambiguity present in real-world data.
Traditional hypergraphs rely on predefined rules (\eg, thresholding based on geometric distances or similarity measures) to form hyperedges, leading to hard, binary memberships. This explicit and rigid formulation fails to capture the pervasive ambiguity present in real-world data.

In contrast, in soft hypergraphs we consider that in every high-order relationship, all vertices participate to a certain degree, thereby forming continuous and differentiable soft hyperedges. This implicit and flexible definition not only better reflects the ambiguity and complexity inherent in real data but also has the potential to represent more abstract and deeper semantic associations.

Formally, soft hypergraphs can be defined as a triplet:
\begin{equation}
    \mathcal{G} = \{\mathcal{V}, \mathcal{E}, A\},
\end{equation}
where $ \mathcal{V} = \{v_1, v_2, \ldots, v_N\} $ is the set of vertices, comprising $N$ vertices. $ \mathcal{E} = \{e_1, e_2, \ldots, e_M\} $ is the set of hyperedges, comprising $M$ hyperedges. Note that in fact, each hyperedge is the set of all vertices. $ A \in [0,1]^{N \times M} $ is the participation matrix between vertices and hyperedges, where $ A_{n,m} $ represents the degree to which vertex $ v_n $ participates in hyperedge $ e_m $, with its values ranging between 0 and 1.

However, to more succinctly and accurately depict the characteristics of soft hypergraphs, we prefer a matrix-style representation:
\begin{equation}
    \mathcal{G} = \{X, A\},
\end{equation}
where $X \in \mathbb{R}^{N \times D}$ is the vertex feature matrix (\ie, a token sequence) with $D$ as the feature dimension. $A \in [0,1]^{N \times M}$ is the participation matrix, which directly characterizes the high-order semantic associations among vertices using continuous and differentiable values, rather than relying on the traditional binary incidence matrix $H \in \{0,1\}^{N \times M}$.

In summary, our proposed soft design offers two significant advantages. On the one hand, continuous and differentiable hyperedges can more delicately describe the relationships between vertices and hyperedges, capturing varying degrees of association. For example, in scenarios with partial occlusion or ambiguity, the relationship between vertices and hyperedges is often not absolute, and soft hypergraphs are better able to reflect such gradual and uncertain interactions. On the other hand, soft hypergraphs enable the dynamic and learnable generation of hyperedges, allowing the model to adaptively capture complex patterns hidden in large-scale data. This characteristic is crucial for constructing efficient and flexible models in practical visual tasks (a detailed discussion will be presented in Section \ref{sec-c}).

\subsection{Overview of SoftHGNN}
\label{sec-b}

Based on the concept of soft hypergraphs, we propose a novel hypergraph computation framework, namely the Soft Hypergraph Neural Network (SoftHGNN), as shown in Figure \ref{fig:framework}. This architecture is designed to dynamically generate soft hyperedges that effectively capture high-order semantic relationships in visual data, addressing the limitations of traditional hypergraphs in terms of hyperedge redundancy and rigid binary associations. 
% Furthermore, SoftHGNN is a plug-and-play module that can be easily inserted into any model for visual recognition tasks and significantly improves the performance of the model. 
\textcolor{reColor}{Furthermore, SoftHGNN is a lightweight add-on module that can be attached to semantically mature stages of existing vision pipelines for visual recognition tasks.}
Overall, SoftHGNN consists of the following three core parts:

\textbf{1) Soft Hyperedge Generation.} 
We leverage the input token sequence as vertices and utilize a set of learnable hyperedge prototype vectors. In addition, a context-aware offset prediction network is employed to generate sample-specific hyperedge prototypes. Each vertex computes its participation weight based on the semantic distance between its feature and the hyperedge prototypes to obtain the participation matrix $A$. This process allows each hyperedge to establish flexible and continuous connections with all vertices, thereby accurately reflecting complex high-order semantic correlations.
    
\textbf{2) Message Passing on Soft Hypergraph.} 
On the constructed soft hypergraph, vertices interact via soft hyperedges through a message passing mechanism that facilitates high-order interactions. Specifically, this involves two stages: aggregation from vertices to soft hyperedges ($\mathcal{V}\rightarrow \mathcal{E}$) and information dissemination from soft hyperedges back to vertices ($\mathcal{E}\rightarrow \mathcal{V}$).
% , thus enabling efficient learning of high-order relationships.
    
\textbf{3) Sparse Soft Hyperedge Selection (Optional).}
Typically, the basic SoftHGNN architecture, composed of modules \textbf{1)} and \textbf{2)}, suffices for most scenarios. However, in some more complex situations, there may exist a larger number of intricate high-order relationships. To expand the number of soft hyperedges without incurring additional computational complexity, we introduce the Sparse Hyperedge Selection (SeS) strategy. Moreover, to prevent over-activation or under-utilization of a few hyperedges, we propose a load-balancing regularizer to encourage all hyperedges to contribute evenly during training.

In the following sections, we will delve into the theoretical details of soft hyperedge generation, message passing on soft hypergraph, and sparse hyperedge selection, respectively.

\subsection{Soft Hyperedge Generation}
\label{sec-c}

The soft hyperedge generation module dynamically constructs a continuous and differentiable participation matrix $A$ based on the input vertex features, enabling each vertex to participate in each hyperedge to varying degrees. This module not only leverages local vertex features but also fully utilizes global contextual information. \textcolor{reColor}{By incorporating a head-wise partition mechanism and pre-projection strategy, it achieves fine-grained capture of high-order semantic relationships.} The generation of soft hyperedges comprises three components: dynamic hyperedge prototype generation, vertex feature pre-projection, and participation matrix generation.

\subsubsection{Dynamic Hyperedge Prototype Generation}

To generate hyperedge prototypes tailored to each sample, we pre-define a set of globally shared learnable hyperedge prototype parameters $P_0 \in \mathbb{R}^{M \times D}$, where $M$ denotes the preset number of hyperedges and $D$ is the feature dimension for each vertex. In combination with the global contextual features of each sample, we generate a sample-specific hyperedge prototype offset $\Delta P$.

Specifically, let the vertex feature matrix of a sample be $X \in \mathbb{R}^{N \times D}$, where $N$ is the number of vertices (\ie, tokens).
% and $D$ is the feature dimension. 
For this sample, we denote the feature representations of the vertices as
$
\{x_1, x_2, \ldots, x_N\}, \quad x_i \in \mathbb{R}^D.
$
To extract global contextual information, we perform average pooling over all vertices to obtain $f_{\text{avg}}$,
% \begin{equation}
%     f_{\text{avg}} = \frac{1}{N} \sum_{i=1}^{N} x_i \in \mathbb{R}^D,
% \end{equation}
and we perform max pooling over each feature dimension to obtain the max-pooled feature $f_{\text{max}}$.
% \begin{equation}
%     f_{\text{max}} = \Bigl( \max_{1 \le i \le N} x_{i,1},\, \max_{1 \le i \le N} x_{i,2},\, \dots,\, \max_{1 \le i \le N} x_{i,D} \Bigr) \in \mathbb{R}^D,
% \end{equation}
% where $x_{i,j}$ denotes the $j$-th feature of the $i$-th vertex. 
By concatenating these two pooled results along the feature dimension, we obtain the global context vector for the sample:
$
        f_{\text{global}} = \begin{pmatrix} f_{\text{avg}} , f_{\text{max}} \end{pmatrix} \in \mathbb{R}^{2D}
$.
Next, a context-aware offset prediction network $\phi$ processes the global context vector to generate the hyperedge prototype offset:
\begin{equation}
    \Delta P = \phi\bigl(f_{\text{global}}\bigr) \in \mathbb{R}^{M \times D}.
\end{equation}
Finally, by adding the global hyperedge prototype $P_0$ to the offset $\Delta P$, we obtain the dynamic hyperedge prototype for the sample:
\begin{equation}
    P = P_0 + \Delta P \in \mathbb{R}^{M \times D}.
\end{equation}
This design ensures that the generated hyperedge prototypes incorporate both the globally shared prior information and the sample-specific contextual adjustments, which is fundamental for accurately capturing high-order semantic relationships.

\subsubsection{Vertex Feature Pre-projection}

\textcolor{reColor}{To enhance semantic representation, we pre-project the vertex features and introduce a head-wise partition mechanism. Specifically, the input vertex features are mapped to a new space via a linear transformation:}
\begin{equation}
    X_{\text{proj}} =  X W_{\text{pre}}\in \mathbb{R}^{N \times D},
\end{equation}
where $W_{\text{pre}} \in \mathbb{R}^{D \times D}$ is the pre-projection parameter matrix. Subsequently, the pre-projected vertex features are split into $h$ heads, with each head having a dimension of $D_{\text{head}} = \frac{D}{h}$.
We denote the head-wise vertex features as 
$X_{\text{heads}} \in \mathbb{R}^{h \times N \times D_{\text{head}}}.$
Similarly, the dynamic hyperedge prototypes $P \in \mathbb{R}^{M \times D}$ are divided into heads, yielding 
$P_{\text{heads}} \in \mathbb{R}^{h \times M \times D_{\text{head}}}.$
\textcolor{reColor}{This head-wise partition decomposes the similarity computation into multiple lower-dimensional groups and then aggregates the resulting scores, reducing reliance on a single full-dimensional similarity estimate for participation matrix generation.}

\subsubsection{Participation Matrix Generation}
\label{sec-433}

After obtaining the head-wise representations of vertex features and hyperedge prototypes, we compute the participation matrix $A$ by measuring the dot-product similarity, which quantifies the degree of participation of each vertex in each hyperedge. For each head $\tau \in \{1,2,\dots,h\}$, the similarity between vertices and hyperedges is computed as:
\begin{equation}
    S^{(\tau)} = \frac{X_{\text{heads}}^{(\tau)} \cdot \Bigl(P_{\text{heads}}^{(\tau)}\Bigr)^\top}{\sqrt{D_{\text{head}}}} \in \mathbb{R}^{N \times M}.
\end{equation}
Here, dividing by $\sqrt{D_{\text{head}}}$ stabilizes the gradients by preventing the dot-product values from growing too large as the dimensionality increases. 
To fuse the information across all heads, we average the similarity scores from each head:
\begin{equation}
    S =  \frac{1}{h} \sum_{\tau=1}^{h} S^{(\tau)} \in \mathbb{R}^{N \times M}.
\end{equation}
This operation produces a single vertex-hyperedge score matrix for the generation of the participation matrix $A$.

Finally, we apply Softmax normalization to the similarity score matrix $S$ to obtain the continuous and differentiable participation matrix $A$. Specifically, we provide two normalization approaches. The first is normalization along the hyperedge dimension (referred to as E-Norm), which is our default approach. Its purpose is to ensure that the sum of the participation scores of all vertices within the same hyperedge equals one. Formally:
\begin{equation}
    A_{i,j} = \frac{\exp(S_{i,j})}{\sum_{k=1}^{N} \exp(S_{k,j})},
\end{equation}
where $i \in \{1, 2, \ldots, N\}$ and $j \in \{1, 2, \ldots, M\}$.
The second method is normalization along the vertex dimension (referred to as V-Norm). It ensures that for a given vertex, the sum of its participation scores across all hyperedges equals 1. Formally:
\begin{equation}
    A_{i,j} = \frac{\exp(S_{i,j})}{\sum_{k=1}^{M} \exp(S_{i,k})}.
\end{equation}

% % Finally, we apply the Softmax function to normalize the similarity vector for each vertex, yielding the continuous and differentiable participation matrix:
% Finally, we apply Softmax along the vertex dimension for each hyperedge’s node assignment vector, yielding the continuous and differentiable participation matrix:
% \begin{equation}
%     A_{i,j} = \frac{\exp(S_{i,j})}{\sum_{k=1}^{N} \exp(S_{k,j})},
% \end{equation}
% where $i \in \{ 1,2,\ldots,N\}$ and $j \in \{1,2,\ldots,M\}$.

% This dot-product similarity computation allows each vertex to participate in each hyperedge to varying degrees. 
Such a soft allocation mechanism enables the model to finely adjust the contribution of each vertex to the hyperedges, especially in complex semantic scenarios where ambiguity or partial overlap may occur.

\subsection{Message Passing on Soft Hypergraph}
\label{sec-d}

The SoftHGNN leverages the soft hyperedge structure to enable effective interactions between vertices. The general message passing is divided into two stages: \textbf{Aggregation from Vertices to Soft Hyperedges} and \textbf{Dissemination from Soft Hyperedges to Vertices}.

\subsubsection{Aggregation from Vertices to Soft Hyperedges}

The objective of this stage is to collect information from all vertices within a soft hyperedge to form a high-order semantic representation of that soft hyperedge. Consider a soft hyperedge $e_m$, which should aggregate the semantic information from all vertices. Unlike traditional hypergraphs, in SoftHGNN each vertex $v_i$ contributes to the soft hyperedge $e_m$ in a continuous and differentiable manner, as indicated by $A_{i,m} \in [0,1]$.

Thus, the feature representation $f_m$ of soft hyperedge $e_m$ is defined as:
\begin{equation}
    f_m = \sum_{i=1}^{N} A_{i,m} \cdot x_i,
\end{equation}
where $x_i \in \mathbb{R}^D$ represents the feature of vertex $v_i$, and $A_{i,m}$ denotes its participation degree in soft hyperedge $e_m$. This design implies that every vertex contributes to each soft hyperedge with the degree of contribution controlled by $A$. It embodies a soft, high-order aggregation mechanism that can model complex and non-binary semantic relationships found in real-world scenarios.
To further enhance the representation, a linear projection and a non-linear activation are applied to the aggregated soft hyperedge feature:
\begin{equation}
    f_m' = \sigma(W_e f_m),
\end{equation}
where $W_e \in \mathbb{R}^{D' \times D}$ is the transformation weight matrix for soft hyperedges, and $\sigma(\cdot)$ denotes an activation function.

This formulation ensures that the soft hyperedge representations capture rich, high-order semantics through a flexible and adaptive aggregation of vertex information.

\subsubsection{Dissemination from Soft Hyperedges to Vertices}

In this stage, our goal is to propagate the high-order representations of the soft hyperedges back to each vertex, thereby updating the vertex features. For a given vertex $v_i$, we consider the information it receives from all soft hyperedges, and its updated feature is expressed as:
\begin{equation}
   \widetilde x_i = \sum_{m=1}^{M} A_{i,m} \cdot f_m',
\end{equation}
where $f_m'$ is the nonlinearly transformed soft hyperedge feature, and $A_{i,m}$ still represents the participation degree of vertex $v_i$ in soft hyperedge $e_m$. This equation indicates that the update information for each vertex is a weighted aggregation of the information from all soft hyperedges, with each soft hyperedge’s contribution weighted by $A_{i,m}$. Furthermore, the updated vertex feature can be mapped to the target dimension through a linear projection and a non-linear activation:
\begin{equation}
    x_i' = \sigma(W_n \widetilde x_i),
\end{equation}
where $W_n \in \mathbb{R}^{D'' \times D'}$, and $D''$ is the final output dimension.

\subsubsection{Matrix Formulation of Message Passing}

The above two-stage process can be concisely expressed in matrix form as follows.
From vertices to soft hyperedges:
\begin{equation}
      F_e = A^\top X , \quad F_e' = \sigma(F_e W_e^\top) \in \mathbb{R}^{M \times D'}.
\end{equation}
From soft hyperedges to vertices:
\begin{equation}
    \widetilde{X} = AF_e', \quad X'=\sigma(\widetilde{X}W_n^\top) \in \mathbb{R}^{N \times D''}.
\end{equation}
% \begin{equation}
%     X' = \sigma(A F_e' W_n^\top) \in \mathbb{R}^{N \times D''}.
% \end{equation}
Thus, the entire message passing process can be unified into the following compact matrix expression:
\begin{equation}
    X' = \sigma\Bigl( A \, \sigma\bigl( A^\top X\, W_e^\top \bigr) W_n^\top \Bigr).
\end{equation}
This unified expression first aggregates the vertex features into soft hyperedges via $A^\top X\, W_e^\top$, applies a nonlinear mapping to generate soft hyperedge features, and then propagates these features back to the vertices using $A$, followed by an additional linear transformation and nonlinear activation to update the vertex features. Through the continuous participation matrix $A$ and the bidirectional message passing mechanism, SoftHGNN dynamically models high-order semantic relationships. Vertex features are enhanced by the global context while still preserving fine-grained local structures. This design avoids the reliance on hard connections in traditional hypergraphs, thereby equipping the model with stronger expressive power and robustness when dealing with ambiguity, overlap, and structural noise in real-world scenarios.

\subsection{Sparse Soft Hyperedge Selection and Load-Balancing}
\label{sec-e}

\textcolor{reColor}{
Typically, the basic SoftHGNN architecture is sufficient for most scenarios. However, in more complex situations (\eg, dense crowds or high-resolution images), the input may contain a richer and more diverse set of high-order relationships, which calls for a larger candidate hyperedge space. A naive increase in the number of soft hyperedges, however, does not necessarily improve performance: it may introduce overlapping or weakly useful hyperedges, cause semantic diffusion across hyperedges, or lead the model to over-rely on a small subset of hyperedges while leaving others under-utilized.}

\textcolor{reColor}{
To address this issue, and inspired by sparsification in Mixture-of-Experts (MoE) \cite{moe_survey,deepseekmoe}, we propose a Sparse Hyperedge Selection (SeS) strategy combined with load-balancing regularization. The goal is not only to control computation, but also to encourage specialization among candidate hyperedges when the hyperedge capacity is expanded. Specifically, we employ a larger set of candidate hyperedges to cover richer high-order associations, while sparsely selecting only the most important ones for message passing and feature updating. Meanwhile, the load-balancing regularizer discourages the model from over-activating only a few hyperedges and promotes more balanced utilization across the candidate set. In the following, we describe the sparse hyperedge selection strategy and the load-balancing regularizer, respectively.}

\subsubsection{Sparse Hyperedge Selection Strategy}

In the soft hyperedge generation module, assume that $M$ hyperedges are obtained, with the initial (unnormalized) participation matrix given by $S \in \mathbb{R}^{N \times M}$.
Note that $S$ denotes the unnormalized similarity matrix before applying Softmax (see Section \ref{sec-433}).
We partition the soft hyperedge set $\mathcal{E}$ into two subsets:
% \begin{equation}
%     \mathcal{E} = \mathcal{E}^\text{fixed} \cup \mathcal{E}^\text{dyn}, \quad |\mathcal{E}^\text{fixed}| = M_\text{fixed}, \quad |\mathca l{E}^\text{dyn}| = M_\text{dyn}, \\
%     \mathrm{ s.t. }
%     M = M_\text{fixed} + M_\text{dyn}.
% \end{equation}
\begin{equation}
\small
  \begin{gathered}
    \mathcal{E}
    = \mathcal{E}^{\mathrm{fixed}} \cup \mathcal{E}^{\mathrm{dyn}},
    ~
    \bigl|\mathcal{E}^{\mathrm{fixed}}\bigr| = M_{\mathrm{fixed}},
    ~
    \bigl|\mathcal{E}^{\mathrm{dyn}}\bigr| = M_{\mathrm{dyn}},\\
    \mathrm{s.t.\ }M = M_{\mathrm{fixed}} + M_{\mathrm{dyn}}.
  \end{gathered}
\end{equation}
The fixed soft hyperedges are always active, whereas the dynamic soft hyperedges are subject to sparse selection. For each dynamic soft hyperedge, we compute its global activation score by summing its assignment values over all vertices:
\begin{equation}
    g_j = \sum_{i=1}^{N} S_{i,j}^\text{dyn}, \quad j = 1, \dots, M_{\text{dyn}},
\end{equation}
where $S^\text{dyn}$ denotes the portion of the initial participation matrix $S$ corresponding to the $M_{\text{dyn}}$ dynamic soft hyperedges.
We then rank the dynamic hyperedges by their global activation scores $\{g_j\}$ and select the Top-$k$ ones.
Theoretically, this step can be viewed as selecting a subset $\mathcal{E}^\text{sel} \subset \mathcal{E}^\text{dyn}$ with $|\mathcal{E}^\text{sel}| = k$.
Finally, the hyperedge set is updated as
$
    \mathcal{E} = \mathcal{E}^\text{fixed} \cup \mathcal{E}^\text{sel},
$
and the updated unnormalized participation matrix is formed by the concatenation of these hyperedges' scores.
As a result, the corresponding updated unnormalized participation matrix becomes
$
    S' \in \mathbb{R}^{N \times (M_\text{fixed}+k)}.
$
Then, by applying Softmax normalization to the $S'$, we obtain the final continuous and differentiable participation matrix used for message passing:
\begin{equation}
    A_{i,j}^\text{total} = \frac{\exp(S_{i,j}')}{\sum_{q=1}^N \exp(S_{q,j}')},
\end{equation}
where $i=1,\dots,N$, $ j=1,\dots,(M_\text{fixed}+k)$.

\subsubsection{Load-Balancing Regularizer}

To avoid situations where some soft hyperedges are over-activated or remain largely inactive, we introduce a load-balancing regularizer. 

This design is analogous to the auxiliary load-balancing losses used in MoE-style sparse activation to prevent expert collapse, but here it is applied to the selection of dynamic soft hyperedges and avoids unbalanced use of these soft hyperedges.
During training, we record the selection outcomes of the dynamic soft hyperedges over multiple forward passes. Specifically, suppose that over $T$ forward passes we construct a binary mask $\mathcal{M} \in \mathbb{R}^{T \times M_\text{dyn}}$,
where each entry is defined as
\begin{equation}
    \mathcal{M}_{j}^{(t)} = 
\begin{cases}
1, \text{if dynamic soft hyperedge } e_j \text{ selected}, \\
0, \text{otherwise}.
\end{cases}
\end{equation}
After $T$ forward passes, for each dynamic soft hyperedge we compute its activation probability:
\begin{equation}
    p_j = \frac{1}{T} \sum_{t=1}^{T} \mathcal{M}^{(t)}_j, \quad j = 1, \dots, M_{\text{dyn}}.
\end{equation}
We define the target activation probability, assuming a uniform distribution, as
$
    p_\text{target} = \frac{k}{M_{\text{dyn}}}.
$
Our goal is for the actual activation probability distribution of each dynamic soft hyperedge to approach this ideal uniform distribution. Accordingly, the load-balancing regularization loss is defined as
\begin{equation}
    \mathcal{L}_{\text{LB}} = \frac{1}{M_{\text{dyn}}} \sum_{j=1}^{M_{\text{dyn}}} \left( p_j - p_\text{target} \right)^2.
\end{equation}
% This regularization term encourages the dynamic soft hyperedges to be selected uniformly over multiple forward passes, ensuring that each dynamic hyperedge contributes meaningfully to the message passing process rather than relying solely on a few hyperedges, thereby enhancing the model's robustness as well as the effective utilization of hyperedges.
\textcolor{reColor}{This regularization term encourages the dynamic soft hyperedges to be selected more uniformly over multiple forward passes, ensuring that each dynamic hyperedge contributes meaningfully to the message passing process rather than relying disproportionately on only a few hyperedges. Beyond balanced utilization, this also promotes functional specialization among candidate hyperedges and mitigates hyperedge collapse when the candidate hyperedge space is enlarged.}

\subsection{Complexity Analysis of SoftHGNN}
\label{sec-f}

To better understand the scalability and efficiency of SoftHGNN in practical applications, we analyze its computational complexity from a theoretical perspective.

\subsubsection{The Stage of Soft Hyperedge Generation}
In the soft hyperedge generation module, the vertex feature matrix is given by $X \in \mathbb{R}^{N \times D}$, where $N$ denotes the number of vertices and $D$ is the feature dimension, and the number of soft hyperedges is $M$. We analyze the operations separately:

\textbf{Hyperedge Prototype Offset Prediction}. Generating the hyperedge prototype offset requires one or several linear projections. The complexity is primarily $\mathcal{O}(N \times D)$.
    
\textbf{Head-wise Projection and Similarity Calculation}. After mapping the vertex features and hyperedge prototypes into a head-wise subspace, the dot-product similarity between vertices and the hyperedge prototype is computed. The dot-product similarity calculation for vertices and the prototype requires about $\mathcal{O}(N \times M \times (D/h))$ multiplications per head, and after combining all heads, the complexity is $\mathcal{O}(h \times N \times M \times (D/h)) = \mathcal{O}(N \times M \times D)$.

Thus, the main computational cost in the soft hyperedge generation stage arises from the head-wise projection and similarity calculation, and overall the time complexity for this stage is approximately $\mathcal{O}(N \times M \times D)$.

\subsubsection{The Stage of Message Passing}

The message passing process in SoftHGNN is divided into two stages: Aggregation from Vertices to Soft Hyperedges and Dissemination from Soft Hyperedges to Vertices. We analyze these two stages separately:

\textbf{Aggregation from Vertices to Soft Hyperedges}. If the number of soft hyperedges is $M$ and the number of vertices is $N$, then multiplying $X \in \mathbb{R}^{N \times D}$ with $A^\top \in \mathbb{R}^{M \times N}$ to obtain the hyperedge feature matrix has a cost of $\mathcal{O}(N \times M \times D)$.

\textbf{Dissemination from Soft Hyperedges to Vertices}. Multiplying the obtained hyperedge features by $A \in \mathbb{R}^{N \times M}$ also primarily depends on the dimensions $N \times M$ and the feature dimension $D$. Thus, the cost of this stage is similarly approximated as $\mathcal{O}(N \times M \times D)$.

Therefore, the primary computational complexity for message passing is $\mathcal{O}(N \times M \times D)$.

\subsubsection{Overall Discussion}

Combining the soft hyperedge generation and message passing stages, the overall computational complexity of SoftHGNN is approximately $\mathcal{O}(N \times M \times D)$. Based on empirical observations, the number of hyperedges $M$ is usually a small constant (\eg, 8, 16, \etc), and $D$ is also a fixed constant. Consequently, the complexity of the SoftHGNN architecture can be viewed as linear with respect to the number of vertices (tokens), \ie, $\mathcal{O}(N)$. In contrast to the self-attention architecture and the traditional HGNN architectures discussed earlier, which exhibit quadratic complexity $\mathcal{O}(N^2)$, our SoftHGNN demonstrates a significant computational efficiency advantage.

\section{Experimental Results}

\subsection{Tasks and Datasets}

To validate the superiority and universality of the proposed SoftHGNN, we conduct extensive experiments on three typical visual recognition tasks: image classification, crowd counting, and object detection. In these tasks, we employ a total of five widely-used datasets. In the following, we briefly introduce these datasets by task.

\textbf{Image Classification.}
For the image classification task, we adopt the ImageNet-100 \cite{imgnet} and ImageNet-1k \cite{imgnet} datasets. ImageNet-1k is a widely used large-scale benchmark derived from the ILSVRC 2012 challenge, covering 1,000 diverse object categories. It consists of approximately 1.28 million images for training and 50,000 images for validation. ImageNet-100 is a subset constructed by selecting 100 classes from the original ImageNet-1k dataset. It comprises roughly 130,000 training images and 5,000 validation images, serving as a lighter benchmark to evaluate model performance on a smaller scale.

% \textbf{Image Classification.}
% For the image classification task, we adopted the ImageNet-100 \cite{imgnet} and ImageNet-1k \cite{imgnet} datasets, which are widely used in both academia and industry to evaluate model performance on color image classification. The CIFAR-10 dataset consists of 60,000 images divided into 10 classes. Each class contains 6,000 images, with 5,000 images for training and 1,000 images for testing. The CIFAR-100 is an extension of CIFAR-10, containing 60,000 images divided into 100 classes. Each class comprises 600 images, with 500 images for training and 100 images for testing.

\textbf{Crowd Counting.}
The crowd counting task focuses on accurately estimating the number of targets (persons) in dense scenes. We utilized the ShanghaiTech Part-A and Part-B datasets \cite{shanghaitech}. In each image, every target is annotated with a point near the center of the head, resulting in over 330,000 annotations in total. ShanghaiTech Part-A: contains 482 images, primarily featuring dense crowd scenes, with 300 images in the training set and 182 images in the test set. ShanghaiTech Part-B comprises 716 images with relatively sparser crowd scenes, where 400 images are used for training and 316 images for testing.

\textbf{Object Detection.}
The object detection task requires the model to not only recognize the categories of objects in images but also to accurately localize them. For this purpose, we conducted experiments on the MS COCO dataset \cite{mscoco}. The MS COCO dataset is a large-scale benchmark for object detection, containing over 328,000 images. It provides bounding boxes and instance segmentation masks for 80 object categories, along with extensive keypoint annotations, making it an essential benchmark in the field of object detection.

\begin{figure}[t]
    \centering
    \includegraphics[width=\linewidth]{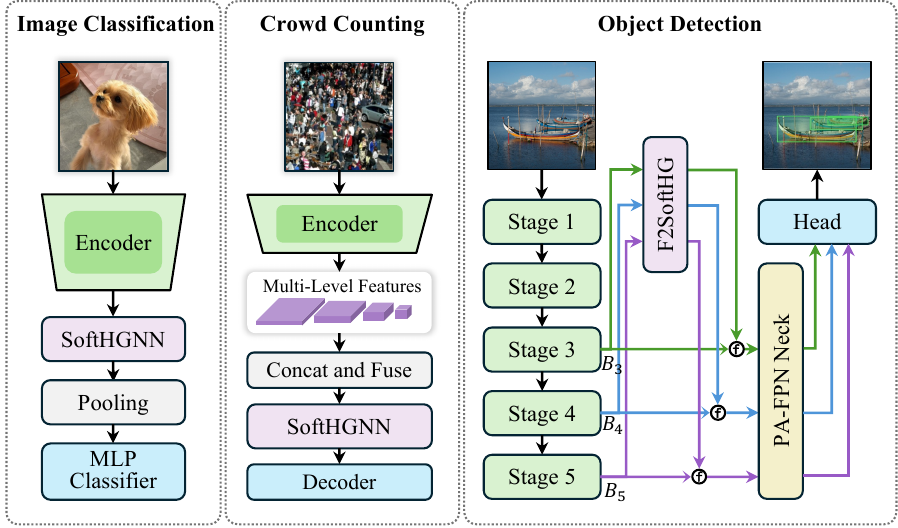}
    \caption{Model settings in image classification, crowd counting, and object detection tasks.}
    \label{fig:tasks}
    % \vspace{-10pt}
\end{figure}

\subsection{Evaluation Metrics}
For the image classification task, we use Top-1 and Top-5 accuracy as the primary evaluation metric, following mainstream works \cite{swin,pvt,vig}.

In the crowd counting task, we use Mean Absolute Error (MAE) and Mean Squared Error (MSE) as evaluation metrics, following previous studies \cite{cctrans,caapn, lsc_cnn}.

% MAE is defined as:
% \begin{equation}
%     \text{MAE} = \frac{1}{N} \sum_{i=1}^{N} \left| \hat{y}_i - y_i \right|,
% \end{equation}
% where $\hat{y}_i$ is the predicted count, $y_i$ is the ground truth count, and $N$ is the total number of images. MSE is defined as:
% \begin{equation}
%     \text{MSE} = \sqrt{ \frac{1}{N} \sum_{i=1}^{N} \left( \hat{y}_i - y_i \right)^2 }.
% \end{equation}

For the object detection task, the model is required to both recognize the object classes and accurately localize their positions in the image. To comprehensively evaluate detection performance, we employ several metrics to quantify both detection accuracy and computational efficiency. The primary accuracy metrics are $\text{AP}_{50}$ and $\text{AP}_{50:95}$, where $\text{AP}_{50}$ denotes the Average Precision when the Intersection over Union (IoU) threshold is set to 0.5, and $\text{AP}_{50:95}$ is the mean Average Precision computed over multiple IoU thresholds (typically from 0.5 to 0.95 in increments of 0.05). 
% In addition to these accuracy metrics, we also calculate FLOPS (Floating Point Operations) to assess the computational complexity during inference. A lower FLOPS value indicates that the model achieves high detection performance while maintaining greater computational efficiency, making it suitable for real-time or resource-constrained applications.

\subsection{Experimental Setup}

\begin{table}[t] % 如果需要跨栏显示，请将 {table} 改为 {table*}
  \centering
  \caption{Comparison of classification performance (\%) with different models on ImageNet-100.}
  \label{tab:cls_100}
  \setlength{\tabcolsep}{3pt}{
    \rowcolors{1}{gray!15}{white}
    \begin{tabular}{lccc}
      \hiderowcolors
      \toprule
      \textbf{Method} & \textbf{Top-1} & \textbf{Top-5} & \textbf{Param.} \\
      \midrule
      \showrowcolors
      
      % --- Part 1: Baselines ---
      ResNet-18 \cite{resnet} & 80.7 & 94.7 & 11.2\,M \\
      ViHGNN-Tiny \cite{vihgnn} & 85.4 & 96.9 & 7.3\,M \\
      PVT-Small \cite{pvt} & 84.2 & 96.6 & 24.0\,M \\
      \midrule
      
      % --- Ablation on PVT-Tiny (baseline) ---
      PVT-Tiny & 82.7 & 95.4 & 12.8\,M \\
      +Attention \cite{vit} & 83.0 \textcolor{deepgreen}{($\uparrow$0.3)} & 95.7 \textcolor{deepgreen}{($\uparrow$0.3)} & 14.6\,M \\
      +HGNN($\epsilon$-ball) \cite{hgnnp} & 83.1 \textcolor{deepgreen}{($\uparrow$0.4)} & 96.3 \textcolor{deepgreen}{($\uparrow$0.9)} & 14.0\,M \\
      +HGNN($k$-NN) \cite{hgnnp} & 82.9 \textcolor{deepgreen}{($\uparrow$0.2)} & 95.2 \textcolor{deepred}{($\downarrow$0.2)} & 14.0\,M \\
      +HgVT Block \cite{hgvt} & 85.1 \textcolor{deepgreen}{($\uparrow$2.4)} & 96.5 \textcolor{deepgreen}{($\uparrow$1.1)} & 14.7\,M \\
      \textbf{+SoftHGNN} & \textbf{86.1 \textcolor{deepgreen}{($\uparrow$3.4)}} & \textbf{97.2 \textcolor{deepgreen}{($\uparrow$1.8)}} & 14.2\,M \\
      \textbf{+SoftHGNN-SeS} & \textbf{86.8 \textcolor{deepgreen}{($\uparrow$4.1)}} & \textbf{98.0 \textcolor{deepgreen}{($\uparrow$2.6)}} & 16.4\,M \\
      \midrule
      
      % --- Ablation on SwinT-Tiny (baseline) ---
      SwinT-Tiny \cite{swin} & 88.0 & 97.6 & 27.6\,M \\
      +Attention \cite{vit} & 89.3 \textcolor{deepgreen}{($\uparrow$1.3)} & 98.0 \textcolor{deepgreen}{($\uparrow$0.4)} & 30.0\,M \\
      +HGNN($\epsilon$-ball) \cite{hgnnp} & 89.2 \textcolor{deepgreen}{($\uparrow$1.2)} & 98.0 \textcolor{deepgreen}{($\uparrow$0.4)} & 28.5\,M \\
      +HGNN($k$-NN) \cite{hgnnp} & 89.0 \textcolor{deepgreen}{($\uparrow$1.0)} & 98.0 \textcolor{deepgreen}{($\uparrow$0.4)} & 28.5\,M \\
      +HgVT Block \cite{hgvt} & 89.4 \textcolor{deepgreen}{($\uparrow$1.4)} & 97.8 \textcolor{deepgreen}{($\uparrow$0.2)} & 28.8\,M \\
      \textbf{+SoftHGNN} & \textbf{90.1 \textcolor{deepgreen}{($\uparrow$2.1)}} & \textbf{98.3 \textcolor{deepgreen}{($\uparrow$0.7)}} & 28.6\,M \\
      \textbf{+SoftHGNN-SeS} & \textbf{90.9 \textcolor{deepgreen}{($\uparrow$2.9)}} & \textbf{98.7 \textcolor{deepgreen}{($\uparrow$1.1)}} & 30.7\,M \\
      
      \bottomrule
    \end{tabular}
  }
\end{table}

Experiments are conducted on NVIDIA RTX 4090 GPUs. 
% Specifically, we use a single GPU for image classification and crowd counting tasks, while object detection is trained in parallel using 8 RTX 4090 GPUs. 
For the image classification task, we adopt Pyramid Vision Transformer and Swin-Transformer as the baseline methods. On both ImageNet-100 and ImageNet-1k, we set the learning rate to \(1\times 10^{-3}\) and use the Adam optimizer \cite{adam} with a cosine annealing learning rate scheduler. Standard data augmentation techniques, including random cropping, random scaling, and horizontal flipping, are applied during training. On ImageNet-100, the batch size is set to 512, and the epoch number is set to 200. On ImageNet-1k, the batch size is set to 1024, and the epoch number is set to 300.
For the crowd counting task, we adopt CCTrans \cite{cctrans} as our baseline method, which is a representative Transformer-based crowd counting framework. We strictly follow the original experimental setup of CCTrans to ensure fairness in comparison.
For the object detection task, we select state-of-the-art detection models YOLO11 \cite{yolo11} and YOLOv12 \cite{yolov12} as baselines. We faithfully reproduce and evaluate these models under their original experimental settings \cite{yolo11,yolov12} to ensure a fair and rigorous comparison. When testing the latency and FLOPs of the model, we utilized one Tesla T4 GPU with TensorRT FP16, which is the same as the official settings for YOLO11 and YOLOv12.

\textcolor{reColor}{Unless otherwise specified, all models are initialized from random weights and trained end-to-end.}
By default, both the number of heads and the number of soft hyperedges in SoftHGNN are set to 8. In the variant with Soft Hyperedge Selection, referred to as SoftHGNN-SeS, the number of fixed soft hyperedges is set to 16, with an additional 32 dynamic soft hyperedges from which a sparse subset is selected. A sparsity ratio of 50\% is applied, meaning that 16 out of the 32 dynamic soft hyperedges are activated during each forward pass. When using SoftHGNN-SeS, the final loss is computed as the sum of load-balancing regularization loss and the original loss of the baseline.

\begin{table}[t]
  \centering
  % 请根据实际使用的数据集修改标题
  \caption{Comparison of classification performance (\%) with different models on ImageNet-1k.}
  \label{tab:cls_1k}
  \setlength{\tabcolsep}{3pt}{
    \rowcolors{1}{gray!15}{white}
    \begin{tabular}{lccc}
      \hiderowcolors
      \toprule
      \textbf{Method} & \textbf{Top-1} & \textbf{Top-5} & \textbf{Param.} \\
      \midrule
      \showrowcolors
      % --- Part 1: SOTA Methods ---
      ResNet-18 \cite{resnet} & 70.6 & 89.7 & 11.7\,M \\
      ResNet-50 \cite{resnet} & 75.3 & 92.2 & 25.6\,M \\
      ViG \cite{vig} & 73.9 & 92.0 & 7.1\,M \\
      ViHGNN \cite{vihgnn} & 74.3 & 92.5 & 8.2\,M \\
      \midrule % 分割线，区分 SOTA 对比与消融实验
      % --- Part 2: Ablation / Variants ---
      PVT-Tiny \cite{pvt} & 74.8 & 92.3 & 13.2\,M \\
        +Attention \cite{vit} & 75.1 \textcolor{deepgreen}{($\uparrow$0.3)} & 92.6 \textcolor{deepgreen}{($\uparrow$0.3)} & 15.1\,M \\
        +HGNN($\epsilon$-ball) \cite{hgnnp} & 75.0 \textcolor{deepgreen}{($\uparrow$0.2)} & 92.4 \textcolor{deepgreen}{($\uparrow$0.1)} & 14.5\,M \\
        +HGNN($k$-NN) \cite{hgnnp} & 74.9 \textcolor{deepgreen}{($\uparrow$0.1)} & 92.4 \textcolor{deepgreen}{($\uparrow$0.1)} & 14.5\,M \\
        +HgVT Block \cite{hgvt} & 75.1 \textcolor{deepgreen}{($\uparrow$0.3)} & 92.5 \textcolor{deepgreen}{($\uparrow$0.2)} & 15.2\,M \\
        \textbf{+SoftHGNN} & \textbf{75.6 \textcolor{deepgreen}{($\uparrow$0.8)}} & \textbf{92.9 \textcolor{deepgreen}{($\uparrow$0.6)}} & 14.7\,M \\
        \textbf{+SoftHGNN-SeS} & \textbf{75.9 \textcolor{deepgreen}{($\uparrow$1.1)}} & \textbf{93.1 \textcolor{deepgreen}{($\uparrow$0.8)}} & 16.9\,M \\
      \bottomrule
    \end{tabular}
  }
\end{table}

\begin{table*}[t]
  \centering
  \small
  \caption{Quantitative result of counting performance on ShanghaiTech Part-A and Part-B datasets. }

\resizebox{\textwidth}{!}{
  
  \setlength{\tabcolsep}{1mm}{
  
  \rowcolors{1}{gray!15}{white}

\begin{threeparttable}[b]

  \begin{tabular}{l c  c c c c c}
    \hiderowcolors
        \toprule
        \multirow{2}{*}{\textbf{Method}} & 
        \multirow{2}{*}{\textbf{Backbone}} & 
        \multicolumn{2}{c}{\textbf{SHTech-Part A}} &
        \multicolumn{2}{c}{\textbf{SHTech-Part B}} &
        \multirow{2}{*}{\textbf{Param.}} \\
        \noalign{\vskip -2pt}
        \cmidrule(lr){3-4}\cmidrule(lr){5-6}
          &    & MAE ($\downarrow$) & MSE ($\downarrow$) & MAE ($\downarrow$) & MSE ($\downarrow$) & \\
    \showrowcolors
    \midrule
    CSRNet \cite{csrnet}          & VGG-16 \cite{vgg}           & 68.2 & 115.0 & 10.6 & 16.0  & 16.3 M \\
    BL \cite{bl}  & VGG-19 \cite{vgg} & 62.8 & 101.8 & 7.7 & 12.7 & 21.5 M \\
    DM-Count \cite{dmcount} & VGG-19 \cite{vgg} & 59.7 & 95.7 & 7.4 & 11.8 & 21.5 M \\
    % KDMG \cite{kdmg}           & TPAMI-2020 & CSRNet \cite{csrnet}           & 63.8 & 99.2  & 7.8  & 12.7  & 16.3 \\
    TopoCount \cite{topocount}      & VGG-16 \cite{vgg}      & 61.2 & 104.6 & 7.8  & 13.7 & 25.8 M \\
    LSC-CNN \cite{lsc_cnn}       & VGG-16 \cite{vgg}           & 66.4 & 117.0 & 8.1  & 12.7  & 35.1 M \\
    Ctrans-MISN \cite{ctrans_misn}      & Swin-Trans \cite{swin} & 55.8 & 95.9  & 7.3  & 11.4  & - \\
    CLTR \cite{cltr}           & ViT \cite{vit}              & 56.9 & 95.2  & 6.5  & 10.6  & 43.5 M \\
    GMS \cite{gms}            & HRNet \cite{hrnet}            & 68.8 & 138.6 & 16.0 & 33.5  & - \\
    GCFL \cite{gcfl}          & VGG-19 \cite{vgg}           & 57.5 & 94.3  & 6.9  & 11.0  & 21.5 M \\
    % DDRANet \cite{ddranet} & SPL-2024 & VGG-16 \cite{vgg} & 52.1 & 78.4 & 6.9 & 10.3 & - \\
    Gramformer \cite{gramformer}      & VGG-19 \cite{vgg}         & 54.7 & 87.1  & -    & -     & 29.1 M \\
    CAAPN \cite{caapn}           & VGG-16 \cite{vgg}          & 54.4 & 97.3  & 5.8  & 9.8   & - \\
    mPrompt \cite{mprompt}         & HRNet \cite{hrnet}           & 53.2 & 85.4  & 6.3  & 9.8   & 48.3 M \\
    \midrule
    CCTrans \cite{cctrans}        & Twins-PCPVT \cite{twins}     & 54.8 & 89.0  & 7.6  & 12.2  & 106.4 M  \\
    + Self-Attn \cite{vit}           & Twins-PCPVT \cite{twins}     & 54.7 \textcolor{deepgreen}{($\downarrow$0.2\%)} & 89.7 \textcolor{deepred}{($\uparrow$0.8\%)}  & 7.4 \textcolor{deepgreen}{($\downarrow$2.6\%)}  & 10.9 \textcolor{deepgreen}{($\downarrow$10.7\%)}  & 107.6 M \\
    + HGNN \cite{hgnnp}               & Twins-PCPVT \cite{twins}     & 53.7 \textcolor{deepgreen}{($\downarrow$2.0\%)} & 82.7 \textcolor{deepgreen}{($\downarrow$7.1\%)}  & 7.2 \textcolor{deepgreen}{($\downarrow$5.3\%)}  & 10.9 \textcolor{deepgreen}{($\downarrow$10.7\%)}  & 108.0 M \\
    \textbf{+ SoftHGNN}     & Twins-PCPVT \cite{twins}   & \textbf{52.8 \textcolor{deepgreen}{($\downarrow$3.6\%)}} & 82.1 \textcolor{deepgreen}{($\downarrow$7.8\%)}  & 7.1 \textcolor{deepgreen}{($\downarrow$6.6\%)}  & \textbf{10.6 \textcolor{deepgreen}{($\downarrow$13.1\%)}}  & 107.6 M  \\
   \textbf{+ SoftHGNN-SeS}     & Twins-PCPVT \cite{twins}     & 53.4 \textcolor{deepgreen}{($\downarrow$2.6\%)} & \textbf{82.0 \textcolor{deepgreen}{($\downarrow$7.9\%)} } & \textbf{7.0 \textcolor{deepgreen}{($\downarrow$7.9\%)}}  & \textbf{10.6 \textcolor{deepgreen}{($\downarrow$13.1\%)}}  & 112.9 M  \\
    \midrule
    CCTrans \cite{cctrans}        & Pyramid ViT \cite{pvtv2}      & 53.7 & 85.9  & 7.4  & 12.5  & 49.9 M  \\
    + Self-Attn \cite{vit}         & Pyramid ViT \cite{pvtv2}      & 53.1 \textcolor{deepgreen}{($\downarrow$1.1\%)} & 83.1 \textcolor{deepgreen}{($\downarrow$3.3\%)}  & 7.1 \textcolor{deepgreen}{($\downarrow$4.1\%)}  & 10.9 \textcolor{deepgreen}{($\downarrow$12.8\%)}  & 51.2 M  \\
    + HGNN \cite{hgnnp}              & Pyramid ViT \cite{pvtv2}     & 53.5 \textcolor{deepgreen}{($\downarrow$0.4\%)} & 85.5 \textcolor{deepgreen}{($\downarrow$0.5\%)}  & 7.0 \textcolor{deepgreen}{($\downarrow$5.4\%)}  & 10.9 \textcolor{deepgreen}{($\downarrow$12.8\%)}  & 51.6 M  \\
    \textbf{+ SoftHGNN}     & Pyramid ViT \cite{pvtv2}    & \textbf{51.7 \textcolor{deepgreen}{($\downarrow$3.7\%)}} & 81.3 \textcolor{deepgreen}{($\downarrow$5.4\%)}  & 6.8 \textcolor{deepgreen}{($\downarrow$8.1\%)}  & 10.7 \textcolor{deepgreen}{($\downarrow$14.4\%)}  & 51.2 M  \\
    \textbf{+ SoftHGNN-SeS}    & Pyramid ViT \cite{pvtv2}   & \textbf{51.7 \textcolor{deepgreen}{($\downarrow$3.7\%)}} & \textbf{79.2 \textcolor{deepgreen}{($\downarrow$7.8\%)}}  & \textbf{6.6 \textcolor{deepgreen}{($\downarrow$10.8\%)}}  & \textbf{10.5 \textcolor{deepgreen}{($\downarrow$16.0\%)}}  & 56.4 M  \\
    \bottomrule
\end{tabular}
  \begin{tablenotes}
    % \footnotesize
    \item The results in green indicate improvements over the baseline, while the results in red indicate a decline compared to the baseline. The bolded results represent the most significant improvement over the baseline.
  \end{tablenotes}
\end{threeparttable}
}
}
\label{tab:crowd_counting}
\end{table*}

% ----------------- nano型号表格 -----------------
\begin{table*}[t]
\caption{Quantitative results of nano size model object detection performance on MS COCO dataset.} 
  \label{tab:comparison_nano}
  \centering
    \setlength{\tabcolsep}{3.6 mm}{
    \rowcolors{1}{gray!15}{white}
    {%
      \begin{threeparttable}[b]
      \begin{tabular}{lccccc}
  \toprule
  \hiderowcolors
  \textbf{Method} & \textbf{$\text{AP}_{50}^{val}$}  & \textbf{$\text{AP}_{50:95}^{val}$}  & \textbf{FLOPs}  & \textbf{Latency}  & \textbf{Param.} \\
  \showrowcolors
  \midrule
  Gold-YOLO-N \cite{goldyolo}                 
    & 55.7 & 39.6 & 12.1\,G & 2.92\,ms & 5.6\,M \\
  Hyper-YOLO-T \cite{hyperyolo}                 
    & 54.5 & 38.5 & 9.6\,G  & 2.5\,ms  & 3.1\,M \\
  % Hyper-YOLO-N \cite{hyperyolo}               
  %  & 58.3 & 41.8 & 11.4\,G & 2.7\,ms  & 4.0\,M \\
  YOLOv8-N \cite{yolov8}                      
    & 52.6 & 37.4 & 8.7\,G  & 1.77\,ms & 3.2\,M \\
  YOLOv9-T \cite{yolov9}                      
    & 53.1 & 38.3 & 7.7\,G  & 2.48\,ms  & 2.0\,M \\
  YOLOv10-N \cite{yolov10}                     
    & 53.8 & 38.5 & 6.7\,G  & 1.84\,ms & 2.3\,M \\
  \midrule
  YOLO11-N \cite{yolo11}                       
    & 55.3 & 39.4 & 6.5\,G  & 1.50\,ms  & 2.6\,M \\
  YOLO11-N$^\dagger$ \cite{yolo11}                
    & 53.6 & 38.2 & 6.5\,G  & 1.59\,ms & 2.6\,M \\
  YOLO11-SoftHGNN-N      
    & \textbf{55.8 \textcolor{deepgreen}{($\uparrow$2.2)}} 
    & \textbf{40.2 \textcolor{deepgreen}{($\uparrow$2.0)}} 
    & 8.0\,G & 1.99\,ms & 3.2\,M \\
  \midrule
  YOLOv12-N \cite{yolov12}                     
    & 55.4 & 40.1 & 6.0\,G  & 1.60\,ms & 2.5\,M \\
  YOLOv12-N$^\dagger$ \cite{yolov12}               
    & 55.1 & 39.6 & 6.0\,G  & 1.63\,ms & 2.5\,M \\
  YOLOv12-SoftHGNN-N    
    & \textbf{57.0 \textcolor{deepgreen}{($\uparrow$1.9)}} 
    & \textbf{41.3 \textcolor{deepgreen}{($\uparrow$1.7)}} 
    & 7.4\,G & 1.99\,ms & 3.1\,M \\
  \bottomrule
\end{tabular}
\begin{tablenotes}
      \item We strictly reproduce and evaluate models under their original official experimental settings to ensure a fair and rigorous comparison. $\dagger$ denotes the reproduced result under the official setting. Improvements over the reproduced baseline are shown in green.
    \end{tablenotes}
\end{threeparttable}
}}
\end{table*}

\subsection{Quantitative Results}

\subsubsection{Image Classification}
\label{sec_5_4_1}

In the image classification task, we incrementally introduce different modules at the end of the encoder for comparison: Self-Attention \cite{vit}, HGNN (construct the hypergraph through $\epsilon$-ball or $k$-NN) \cite{hgnnp}, HgVT block \cite{hgvt}, the proposed SoftHGNN, and SoftHGNN-SeS (with the Sparse Hyperedge Selection), as illustrated in Figure \ref{fig:tasks}.

On ImageNet-100, we adopt Pyramid Vision Transformer (PVT) \cite{pvt} and Swin-Transformer (SwinT) \cite{swin} as our baseline. As shown in Table \ref{tab:cls_100}, since the backbone itself is an Attention architecture, the performance improvement brought by using additional attention modules on a small scale is limited. It only improves the Top-1 accuracy by 0.3\% and 1.3\% on PVT and SwinT, respectively. When introducing standard HGNN modules, neither the version constructing hypergraphs based on $\epsilon$-ball nor the one based on $k$-NN brings significant improvement. In particular, the $k$-NN-based HGNN decreases the Top-5 performance of PVT by 0.2\%.  The version introducing HgVT significantly outperforms the former three, achieving performance gains of 2.4\% and 1.4\% on PVT and SwinT, respectively. However, the version introducing the proposed SoftHGNN has fewer parameters than HgVT but achieves significantly larger performance gains. It achieves Top-1 accuracy improvements of 3.4\% and 2.1\% on PVT and SwinT, respectively. Finally, the version introducing SoftHGNN-SeS further improves the model's representational capability by expanding the hyperedge capacity, increasing the Top-1 accuracy of PVT and SwinT by 4.1\% and 2.9\%, respectively.

On ImageNet-1k, we adopt PVT as the baseline to further verify the applicability of the proposed SoftHGNN on large-scale datasets. As shown in Table \ref{tab:cls_1k}, the improvement achieved by introducing the proposed SoftHGNN is still significantly higher than other methods, improving the Top-1 and Top-5 accuracy by 0.8\% and 0.6\% respectively, reaching 75.6\% and 92.9\%, while introducing only 1.5M parameters. When SoftHGNN-SeS is introduced, the Top-1 and Top-5 accuracy are further improved to 75.9\% and 93.1\%.

\subsubsection{Crowd Counting}

In the crowd counting task, we adopt the Transformer-based CCTrans \cite{cctrans} as our baseline, and introduce Self-Attention \cite{vit}, HGNN \cite{hgnnp}, SoftHGNN, and SoftHGNN-SeS into the multi-scale feature fusion stage, as illustrated in Figure \ref{fig:tasks}. Table \ref{tab:crowd_counting} presents the experimental results on the ShanghaiTech Part-A and Part-B datasets. We conduct experiments with two different backbone networks: one is the Twins-PCPVT \cite{twins} used by the original CCTrans, and the other is the more lightweight and efficient Pyramid Vision Transformer (PVT) \cite{pvt}.
Under both backbones, the proposed SoftHGNN series significantly outperform both Self-Attention and the traditional HGNN. When using Twins-PCPVT as the backbone, SoftHGNN series achieve the largest improvements of 3.6\% in MAE and 7.9\% in MSE on Part-A, and 7.9\% in MAE and 13.1\% in MSE on Part-B. When using PVT as the backbone, the improvements brought by the SoftHGNN series are even more substantial: the MAE and MSE improve by up to 3.7\% and 7.8\% on Part-A, and by up to 10.8\% and 16.0\% on Part-B.
Furthermore, it is worth noting that even with significant improvements, SoftHGNN series still maintains the efficiency of parameters. Specifically, SoftHGNN introduces only about 1M additional parameters, while SoftHGNN-SeS adds approximately 6M. This demonstrates that our proposed SoftHGNN series effectively capture high-order semantic associations that are overlooked by traditional HGNN and Self-Attention, thereby enhancing the model's representational capacity in an efficient manner.

\begin{table*}[t]
  \centering
  \caption{Quantitative results of small size model object detection performance on MS COCO dataset.}
  \label{tab:comparison_small}
    \setlength{\tabcolsep}{3.6 mm}{
  % \begin{threeparttable}
    \rowcolors{1}{gray!15}{white}
    {
    % \begin{threeparttable}
        \begin{tabular}{lccccc}
        \hiderowcolors
          \toprule
          \textbf{Method}                & \textbf{$\text{AP}_{50}^{val}$} & \textbf{$\text{AP}_{50:95}^{val}$} & \textbf{FLOPs} & \textbf{Latency}  & \textbf{Param.} \\          \midrule
          \showrowcolors
          Gold-YOLO-S \cite{goldyolo}           & 62.5 & 45.4 & 46.0\,G & 3.82\,ms & 21.5\,M \\
          Hyper-YOLO-S \cite{hyperyolo}          & 65.1 & 48.0 & 39.0\,G & 4.7\,ms & 14.8\,M  \\
          YOLOv8-S \cite{yolov8}              & 61.8 & 45.0 & 28.6\,G & 2.33\,ms & 11.2\,M \\
          YOLOv9-S \cite{yolov9}              & 63.4 & 46.8 & 26.4\,G & 3.46\,ms  & 7.1\,M  \\
          YOLOv10-S \cite{yolov10}             & 63.0 & 46.3 & 21.6\,G & 2.49\,ms & 7.2\,M  \\           \midrule
          YOLO11-S \cite{yolo11}              & 63.9 & 46.9 & 21.5\,G & 2.50\,ms  & 9.4\,M  \\
          YOLO11-S$^\dagger$ \cite{yolo11}    & 62.2 & 45.6 & 21.5\,G & 2.57\,ms & 9.4\,M  \\
          YOLO11-SoftHGNN-S         & \textbf{63.8 \textcolor{deepgreen}{($\uparrow$1.6)}}& \textbf{47.1 \textcolor{deepgreen}{($\uparrow$1.5)}} & 27.4\,G  & 3.16\,ms    & 12.1\,M \\          \midrule
          YOLOv12-S \cite{yolov12}             & 64.1 & 47.6 & 19.4\,G & 2.42\,ms & 9.1\,M  \\
          YOLOv12-S$^\dagger$ \cite{yolov12}   & 63.8 & 47.0 & 19.4\,G & 2.49\,ms  & 9.1\,M  \\
          YOLOv12-SoftHGNN-S          & \textbf{64.8 \textcolor{deepgreen}{($\uparrow$1.0)}} & \textbf{48.0 \textcolor{deepgreen}{($\uparrow$1.0)}} & 25.2\,G  & 3.17\,ms  & 11.4\,M \\
          \bottomrule
        \end{tabular}
   
        % \end{threeparttable}
    }}
  % \end{threeparttable}
  % \vspace{-10pt}
\end{table*}

\begin{table*}[t]
  \centering
  \caption{Quantitative results of medium size model object detection performance on MS COCO dataset.}
  \label{tab:comparison_medium}
    \setlength{\tabcolsep}{3.6 mm}{
    \rowcolors{1}{gray!15}{white}
    {
    \begin{threeparttable}[b]
      \begin{tabular}{lccccc}
      \hiderowcolors
        \toprule
        \textbf{Method} & \textbf{$\text{AP}_{50}^{val}$} & \textbf{$\text{AP}_{50:95}^{val}$} & \textbf{FLOPs} & \textbf{Latency} & \textbf{Param.} \\
        \showrowcolors
        \midrule
        Gold-YOLO-M \cite{goldyolo}         & 67.0 & 49.8 & 87.5\,G  & 6.38\,ms & 41.3\,M \\
        Hyper-YOLO-M \cite{hyperyolo}       & 69.0 & 52.0 & 103.3\,G & 9.0\,ms  & 33.3\,M \\
        YOLOv8-M \cite{yolov8}           & 67.2 & 50.3 & 78.9\,G  & 5.09\,ms & 25.9\,M \\
        YOLOv9-M \cite{yolov9}          & 68.1 & 51.4 & 76.3\,G  & 6.18\,ms  & 20.0\,M \\
        YOLOv10-M \cite{yolov10}           & 68.1 & 51.1 & 59.1\,G  & 4.74\,ms & 15.4\,M \\        \midrule
        YOLO11-M \cite{yolo11}            & 68.5 & 51.5 & 68.0\,G  & 4.70\,ms & 20.1\,M \\
        YOLO11-M$^\dagger$ \cite{yolo11}  & 67.3 & 50.5 & 68.0\,G  & 5.04\,ms    & 20.1\,M \\
        YOLO11-SoftHGNN-M        & \textbf{68.1 \textcolor{deepgreen}{($\uparrow$0.8)}} & \textbf{51.1 \textcolor{deepgreen}{($\uparrow$0.6)}} & 85.0\,G & 6.07\,ms & 25.1\,M \\        
        YOLO11-SoftHGNN-M$^\star$        & \textbf{68.0 \textcolor{deepgreen}{($\uparrow$0.7)}} & \textbf{51.0 \textcolor{deepgreen}{($\uparrow$0.5)}} & 81.7\,G &5.82\,ms & 22.8\,M \\ 
        \midrule
        YOLOv12-M \cite{yolov12}           & 69.5 & 52.6 & 59.8\,G  & 4.27\,ms & 19.6\,M \\
        YOLOv12-M$^\dagger$ \cite{yolov12} & 68.6 & 51.9 & 59.8\,G  & 4.75\,ms  & 19.6\,M \\
        YOLOv12-SoftHGNN-M        & \textbf{69.2 \textcolor{deepgreen}{($\uparrow$0.6)}} & \textbf{52.2 \textcolor{deepgreen}{($\uparrow$0.3)}} & 80.2\,G & 6.01\,ms & 26.3\,M \\
        YOLOv12-SoftHGNN-M$^\star$        & \textbf{69.1 \textcolor{deepgreen}{($\uparrow$0.5)}} & \textbf{52.2 \textcolor{deepgreen}{($\uparrow$0.3)}} & 73.6\,G & 5.09\,ms & 22.3\,M \\
        \bottomrule
      \end{tabular}
      \begin{tablenotes}
      \item The $\star$ represents that when we input features into the SoftHGNN module, we use $1\times1$ convolution to compress the channel number and then restore it by using symmetric convolution blocks after the output, in order to reduce the number of parameters.
    \end{tablenotes}
\end{threeparttable}
    }}

\end{table*}

\subsubsection{Object Detection}

For the object detection task, we take the state-of-the-art object detection models YOLO11 \cite{yolo11} and YOLOv12 \cite{yolov12} as baselines. We introduce a SoftHGNN-based feature aggregation and distribution module, named Fuse to SoftHGNN (F2SoftHG), into the neck part of the model, as illustrated in Figure \ref{fig:tasks}. The previous representative real-time detectors Gold-YOLO \cite{goldyolo}, Hyper-YOLO \cite{hyperyolo}, YOLOv8-v10 \cite{yolov8,yolov9,yolov10} are also included in the table for comparison.
% F2SoftHG retains the architectural design philosophy of classic feature extraction modules in the YOLO series. 

F2SoftHG first fuses multi-scale features using a convolutional block. Then, the fused feature map is fed into a group of SoftHGNNs to model high-order semantic associations. Finally, the output of F2SoftHG is redistributed to each level and fused with the feature maps of the corresponding levels via concatenation and convolution. In this way, F2SoftHG fully leverages the advantages of both SoftHGNN and convolution to realize high-order semantic association extraction across spatial and scale dimensions, promoting multi-object and object-background interactions at global level. 

\begin{figure*}[t]
    \centering
    \includegraphics[width=\linewidth]{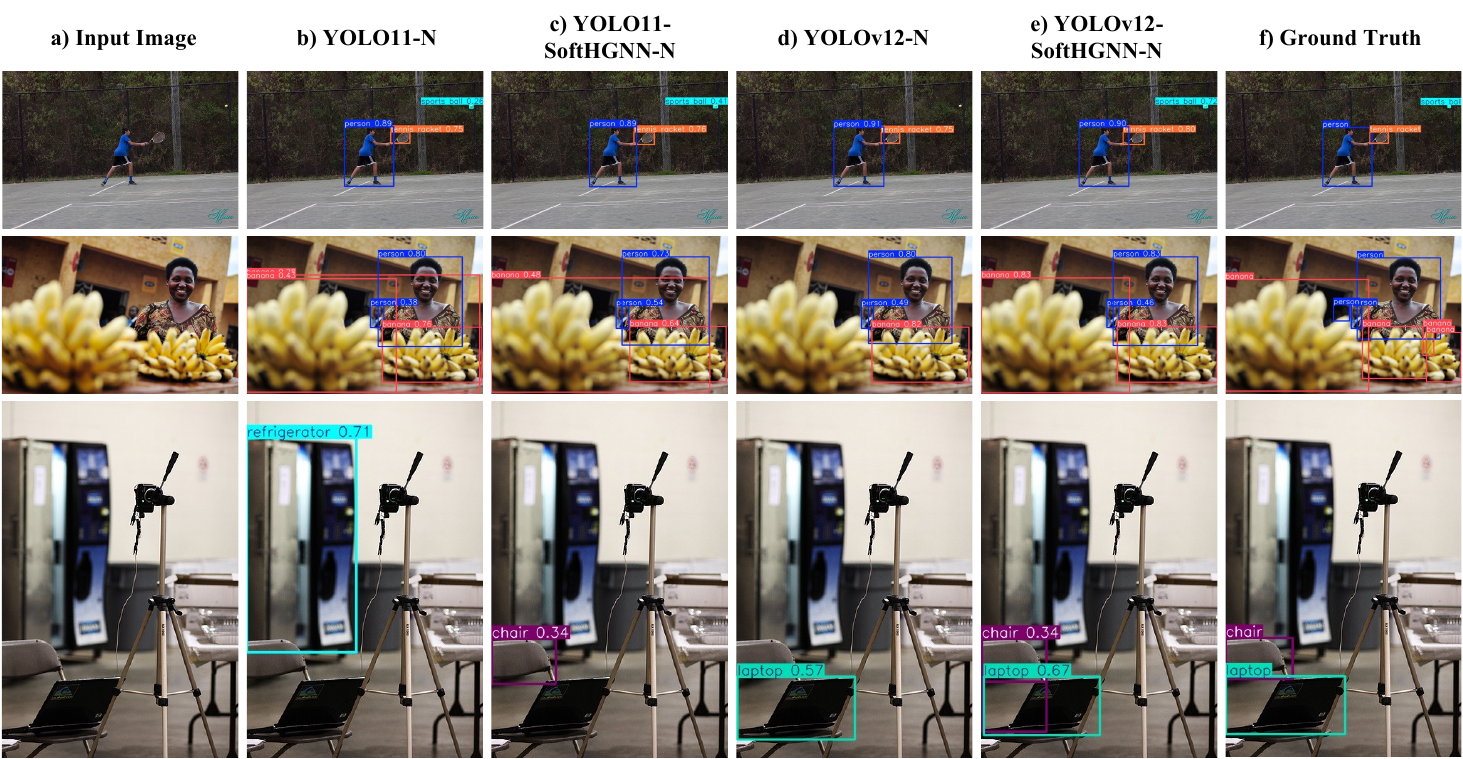}
    \caption{Detection results on the MS COCO dataset. The input image, the prediction of YOLO11-N, YOLO11-SoftHGNN-N, YOLOv12-N, YOLOv12-SoftHGNN-N, and the ground truth are shown from left to right, respectively.}
    \vspace{-8pt} 
    \label{fig:visualize}
\end{figure*}

We conduct experiments on the Nano, Small, and Medium variants of YOLO11 and YOLOv12, as shown in Tables~\ref{tab:comparison_nano}--\ref{tab:comparison_medium}. 
For the Nano scale, our method improves $\text{AP}_{50}^{val}$ by 2.2\% and 1.9\%, and $\text{AP}_{50:95}^{val}$ by 2.0\% and 1.7\% over YOLO11 and YOLOv12, respectively, while introducing only 0.6M additional parameters and about 1.5G extra FLOPs. 
For the Small scale, our method improves $\text{AP}_{50}^{val}$ by 1.6\% and 1.0\%, and $\text{AP}_{50:95}^{val}$ by 1.5\% and 1.0\%. Meanwhile, it still introduces very little computational overhead.
For example, compared with YOLOv12-S, our method introduces an additional 2.3M parameters and 5.8G FLOPs.
For the Medium scale, our method achieves an increase of 0.8\% and 0.6\% in $\text{AP}_{50}^{val}$, and 0.6\% and 0.3\% in $\text{AP}_{50:95}^{val}$ compared to YOLO11 and YOLOv12, respectively. 
Since the multi-scale feature maps in Medium models contain substantially more channels than those in Nano and Small models, directly inserting SoftHGNN leads to a higher overhead (about 5M parameters). 
Thus, we introduce a more parameter-efficient variant (YOLO11-SoftHGNN-M$^\star$ and YOLOv12-SoftHGNN-M$^\star$) by compressing the channel dimension with a $1\times1$ convolution before SoftHGNN and restoring it afterward. 
This lightweight “$\star$” variant reduces the overhead (\ie, to only 2.7M additional parameters), while keeping the accuracy essentially unchanged.
% We conduct experiments on the Nano, Small, and Medium variants of YOLO11 and YOLOv12, as shown in Table \ref{tab:comparison_nano}, Table \ref{tab:comparison_small}, and Table \ref{tab:comparison_medium}. For the Nano scale, our method improves $\text{AP}_{50}^{val}$ by 2.2\% and 1.9\%, and $\text{AP}_{50: 95}^{val}$ by 2.0\% and 1.7\% over YOLO11 and YOLOv12, respectively. Meanwhile, it introduces only an additional 0.6M parameters and about 1.5G FLOPs.
% For the Small scale, our method improves $\text{AP}_{50}^{val}$ by 1.6\% and 1.0\%, and $\text{AP}_{50: 95}^{val}$ by 1.5\% and 1.0\%. Meanwhile, it still introduces only a small number of additional parameters and computational costs. Compared with YOLOv12-S, our method introduces only an additional 2.3M parameters and 5.8G FLOPs, while still maintaining real-time performance and high computational efficiency.
% For the Medium scale, since the multi-scale feature maps contain substantially more channels than those in the Nano and Small models, we additionally introduce a more parameter-efficient version (the YOLO11-SoftHGNN-M$^\star$ and YOLOv12-SoftHGNN-M$^\star$), in which we compress the channel size in the SoftHGNN module. As a result, the normal version achieves an increase of 0.8\% and 0.6\% in $\text{AP}_{50}^{val}$, and 0.6\% and 0.3\% in $\text{AP}_{50: 95}^{val}$ compared to YOLO11 and YOLOv12, respectively. The more lightweight ``$\star$'' version reduces the overhead, introducing only 2.7M parameters, while keeping the accuracy essentially unchanged.

Figure~\ref{fig:visualize} presents the visualization results of YOLO11-N, YOLO11-SoftHGNN-N, YOLOv12-N, and YOLOv12-SoftHGNN-N. From the figure, it can be qualitatively observed that after applying the proposed SoftHGNN, the detection accuracy of YOLO11 and YOLOv12 as well as the confidence of the corresponding objects have been significantly improved.

% \vspace{-15pt}

\subsection{Ablation Study}

\subsubsection{Frozen-Encoder Evaluation}

\textcolor{reColor}{
To examine whether SoftHGNN can also operate effectively as a lightweight add-on module on top of a fixed image encoder, we conduct an additional frozen-encoder study on ImageNet-100. Following the image classification setting in Section~\ref{sec_5_4_1}, the compared modules are attached at the end of the encoder. Specifically, we first train the original backbone on ImageNet-100, and then freeze all encoder parameters. After that, we append SoftHGNN and a newly initialized classification head, and optimize only the newly added module and classifier. For comparison, we also report the standard end-to-end training results under random initialization, as well as the original PVT and SwinT results as the baselines.}

% \begin{table}[t]
%   \centering
%   \caption{\textcolor{reColor}{Ablation study on whether to use a frozen backbone trained on ImageNet-100.}}
%   \label{tab:frozen_ablation}
%   \setlength{\tabcolsep}{2.3 mm}{
%   \begin{tabular}{lccc}
%     \toprule
%     \textbf{Backbone} & \textbf{Frozen} & \textbf{Top-1 (\%)} & \textbf{Top-5 (\%)} \\
%     \midrule
%     PVT-Tiny   & \ding{56} & 86.1 & 97.2 \\
%     PVT-Tiny   & \ding{52}     & 85.5 & 96.7 \\
%     Baseline   & - & 82.7 & 95.4 \\
%     \midrule
%     SwinT-Tiny & \ding{56}  & 90.1 & 98.3 \\
%     SwinT-Tiny & \ding{52}     & 89.5 & 98.1 \\
%     Baseline & - & 88.0 & 97.6 \\
%     \bottomrule
%   \end{tabular}
%   }
% \end{table}

\begin{table}[t]
  \centering
  \caption{\textcolor{reColor}{Ablation study on whether to use a frozen backbone trained on ImageNet-100.}}
  \label{tab:frozen_ablation}
  \setlength{\tabcolsep}{1.7mm}{
  \begin{tabular}{lccc}
    \toprule
    \textbf{Backbone} & \textbf{Enc. Frz.} & \textbf{Top-1 (\%)} & \textbf{Top-5 (\%)} \\
    \midrule
    PVT-Tiny   & -  & 82.7 & 95.4 \\
    +SoftHGNN   & \ding{52}     & 85.5 & 96.7 \\
    +SoftHGNN   & \ding{56} & 86.1 & 97.2 \\
    \midrule
    SwinT-Tiny & - & 88.0 & 97.6 \\
    +SoftHGNN & \ding{52}     & 89.5 & 98.1 \\
    +SoftHGNN & \ding{56}  & 90.1 & 98.3 \\
    \bottomrule
  \end{tabular}
  }
\end{table}

\begin{figure*}[!t]
    \centering
    \includegraphics[width=\linewidth]{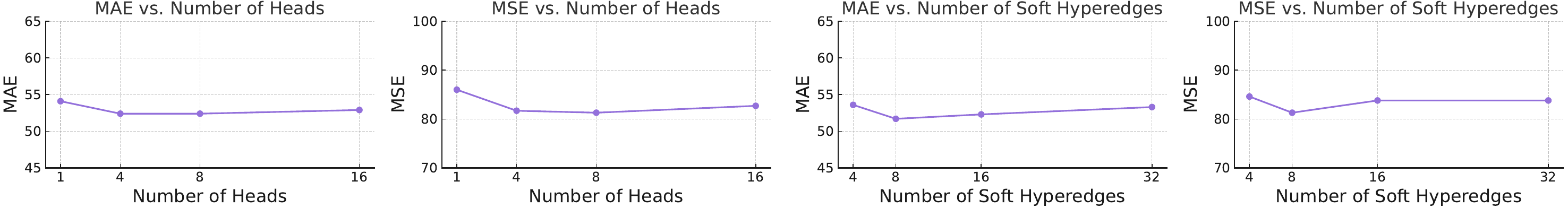}
    \caption{Hyperparameter sensitivity experiment on the ShanghaiTech Part-A dataset.}
    \label{fig:sens}
    % \vspace{-12pt}
\end{figure*}

\textcolor{reColor}{
As shown in Tab.~\ref{tab:frozen_ablation}, SoftHGNN remains effective even when the image encoder is frozen. On PVT-Tiny, the frozen-encoder setting achieves 85.5\% Top-1 and 96.7\% Top-5 accuracy, compared with 86.1\% and 97.2\% under end-to-end training. On SwinT-Tiny, the corresponding results are 89.5\% and 98.1\%, versus 90.1\% and 98.3\% for end-to-end training. Although end-to-end optimization still gives the best performance, the degradation in the frozen setting is small, indicating that SoftHGNN can still serve as an effective lightweight adapter without requiring full backbone retraining.}

\begin{table}[t]
  \centering
  \caption{Comparison of different SoftHGNN variants on Swin and PVT backbones.}
  \label{tab:norm}
  \begin{tabular}{lcc}
    \toprule
    \textbf{Method} & \textbf{Top-1 (\%)} & \textbf{Top-5 (\%)} \\
    \midrule
    SwinT-Tiny          & 88.0 & 97.6 \\
    +SoftHGNN (E-Norm) & \textbf{90.1} & \textbf{98.3} \\
    +SoftHGNN (V-Norm) & 89.8 & 97.9 \\
    +SoftHGNN w/o Norm & 89.4 & 97.4 \\
    \midrule
    PVT-Tiny           & 82.7 & 95.4 \\
    +SoftHGNN (E-Norm) & \textbf{86.1} & \textbf{97.2} \\
    +SoftHGNN (V-Norm) & 85.7 & 97.0 \\
    +SoftHGNN w/o Norm & 85.2 & 96.7 \\
    \bottomrule
  \end{tabular}
\end{table}

\subsubsection{Ablation on Softmax Normalization}

Applying Softmax normalization to the original similarity score matrix $S$ is a crucial step for generating the participation matrix $A$. To investigate the impact of normalization and the two different normalization approaches on the performance of SoftHGNN, we conduct ablation experiments, as shown in Table~\ref{tab:norm}. The results demonstrate that although the model still achieves a substantial improvement over the baseline without Softmax normalization, applying either V-Norm or E-Norm yields significantly better results. This underscores the necessity of the normalization process. Notably, E-Norm slightly outperforms V-Norm; thus, we adopt E-Norm as our default normalization configuration.

\subsubsection{Hyperparameter Sensitivity}

To investigate the impact of different numbers of heads and soft hyperedges on the model performance, we conduct extensive hyperparameter sensitivity experiments on the ShanghaiTech Part-A dataset, as illustrated in Figure \ref{fig:sens}.

First, we fix the number of soft hyperedges to 8 and keep all other experimental settings identical. We evaluate the performance of SoftHGNN under four different head configurations (1, 4, 8, 16). As shown in Figure \ref{fig:sens}, SoftHGNN achieves better MAE and MSE performance when the number of heads is set to 4 or 8, with the best results obtained when using 8 heads.

\begin{figure}[t]
    \centering
    \includegraphics[width=1\linewidth]{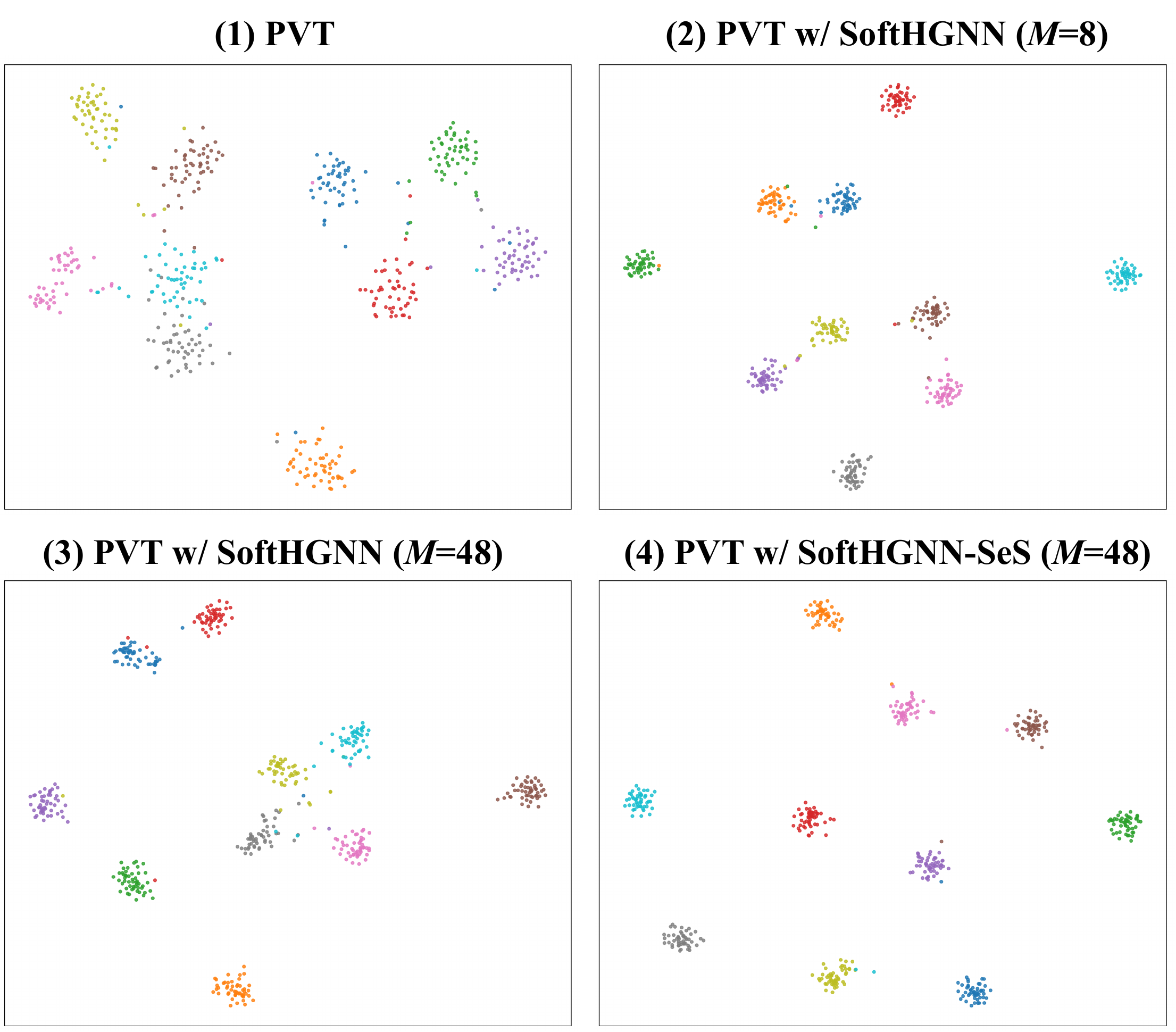}
    \caption{The t-SNE visualizations of the feature representation distributions learned by different model variants on ImageNet-100.}
    \label{fig:tsne}
\end{figure}

\begin{figure*}[t]
    \centering
    \includegraphics[width=1\linewidth]{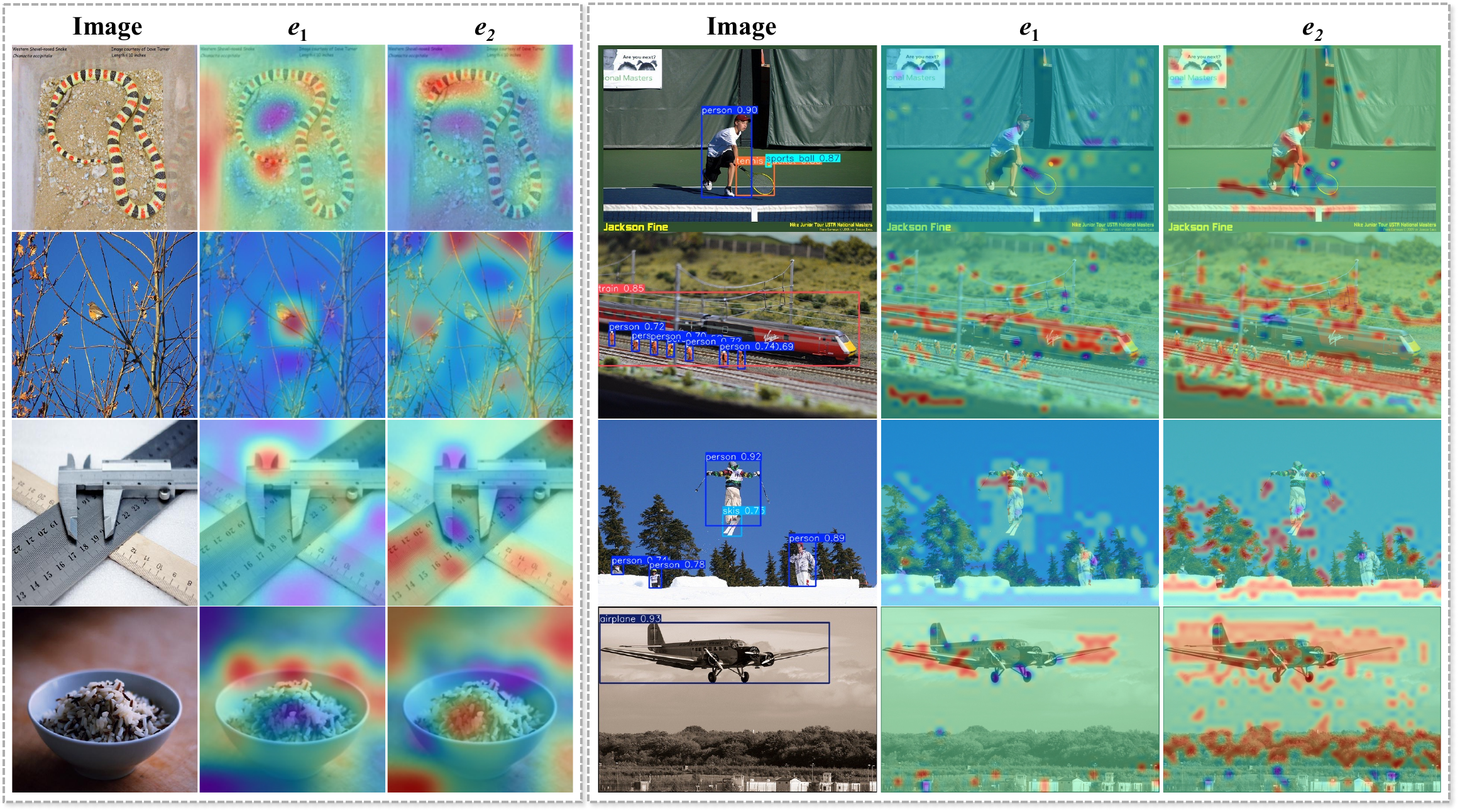}
    \caption{\textcolor{reColor}{Visualization of soft hyperedges on randomly selected validation images for image classification and object detection. The left column displays examples from ImageNet-100 classification, while the right column shows examples from MS COCO object detection.}}
    \label{fig:vis_edge}
\end{figure*}

Next, we fix the number of heads to 8 and maintain all other settings unchanged, while varying the number of soft hyperedges (4, 8, 16, 32). The results show that SoftHGNN performs better when the number of soft hyperedges is set to 8 or 16, with the best MAE and MSE achieved at 8 soft hyperedges. It is noteworthy that in SoftHGNN, increasing the number of soft hyperedges does not always lead to better performance, despite providing a larger capacity for learnable parameters. We attribute this to two reasons. On the one hand, the number of effective high-order semantic relations is directly related to the complexity of the scenes in the dataset. In most cases, a relatively small number of soft hyperedges (such as 8 or 16) is sufficient to capture enough high-order semantic relations. 
% On the other hand, when the number of soft hyperedges becomes too large, the model tends to capture less important or redundant relations, or overfit to specific bad-case associations. 
\textcolor{reColor}{
On the other hand, when the number of soft hyperedges becomes too large, the model may introduce overlapping or weakly useful hyperedges, causing semantic diffusion across hyperedges or over-reliance on a few idiosyncratic associations. SoftHGNN-SeS is exactly proposed to address this issue. By first providing a larger pool of candidate hyperedges and then sparsely selecting only the most important ones, together with load-balancing regularization, SoftHGNN-SeS enables the model to benefit from increased hyperedge capacity while encouraging more specialized and balanced hyperedge utilization.}
% SoftHGNN-SeS is exactly proposed to solve this issue. 
As shown in Table \ref{tab:cls_100}, \ref{tab:cls_1k} and Table \ref{tab:crowd_counting}, SoftHGNN-SeS extends the learning capacity of SoftHGNN by first providing a larger number of dynamic candidate soft hyperedges, and then sparsely selecting only the most important ones. This approach allows the model to benefit from a larger hyperedge space while avoiding redundancy and overfitting, thus significantly enhancing the upper bound of its learning ability.

\subsubsection{Qualitative Analysis}

\begin{figure*}
    \centering
    \includegraphics[width=0.8\linewidth]{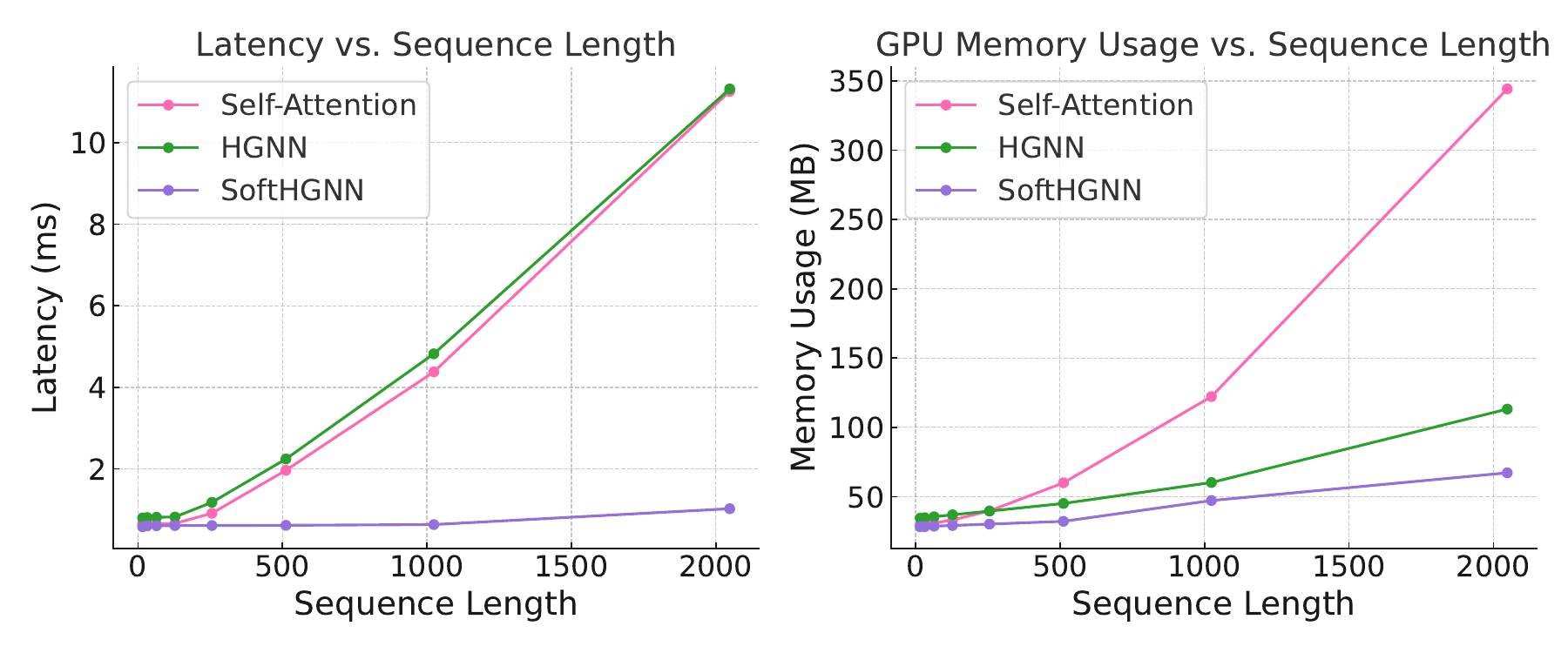}
    \vspace{-5pt}
    \caption{Comparison of latency and GPU memory usage of Self‑Attention, HGNN, and our SoftHGNN under different sequence lengths.}
    \label{fig:latency_mem}
\end{figure*}

We utilize t-SNE to visualize the feature representation distributions learned by different model variants on ImageNet-100, as illustrated in Figure~\ref{fig:tsne}. In the visualization, each point represents the feature representation of a sample, with different colors distinguishing different classes. Ten categories are randomly selected for this visualization. 
Specifically, (1) is the original PVT baseline; (2) is PVT integrated with SoftHGNN, where the number of soft hyperedges is set to 8; (3) also employs SoftHGNN but with the hyperedge count expanded to 48; and (4) is the PVT with SoftHGNN-SeS, with a total of 48 hyperedges (comprising 16 fixed and 32 dynamic hyperedges, with only 16 selected).

Comparing (1) and (2), it is evident that the inclusion of SoftHGNN significantly enhances the separability of feature embeddings across different classes, forming tighter clusters with more distinct boundaries. 
From (2) to (3), as the number of hyperedges increases substantially, the separability between clusters slightly decreases, which may be attributed to the introduction of irrelevant structures for the task during the expansion of hyperedges. 
\textcolor{reColor}{
Finally, comparing (3) and (4), the introduction of the SeS strategy leads to significantly superior cluster separability. This suggests that the SeS strategy helps the model focus on task-relevant relational structures, thereby improving the learned feature representations.}

Furthermore, we visualize soft hyperedges constructed by SoftHGNN in both image classification and object detection tasks, as illustrated in Figure~\ref{fig:vis_edge}. 
The left column displays examples from ImageNet-100 classification, while the right column shows examples from MS COCO object detection. 
In these visualizations, columns $e_1$ and $e_2$ represent two distinct soft hyperedges. Pixels closer to red indicate higher participation weights in the hyperedge, while pixels toward blue indicate lower participation.

These results clearly demonstrate the behavior of SoftHGNN. 
For instance, in the second row of the left column, $e_1$ primarily models the internal correlations within the bird's body, whereas $e_2$ captures the associations between the bird, surrounding branches, and the background sky. 
In the first row of the right column, $e_1$ models the relationships among the tennis ball, the racket, and the athlete's body, while $e_2$ focuses on the interaction between the athlete and the background environment. 
In the second row of the right column, $e_1$ captures the correlations between the train and the nearby staff, while $e_2$ models background elements such as the railway and utility poles.
These visualizations indicate that SoftHGNN effectively learns high-order semantic associations within a scene. 
Moreover, they reveal a form of ``division of labor" through semantic differentiation among different hyperedges. This allows the model to capture diverse relational structures, collectively guiding more accurate decision-making for various visual recognition tasks.

% It is worth noting that HgVT \cite{hgvt} states that these redundant hyperedges may play a certain role in comprehensive visual representation learning. However, for specific tasks, an excessive number of hyperedges may not lead to any performance improvement so that SoftHGNN-SeS is 

\subsubsection{Computational Efficiency}

We compare the latency and GPU memory usage of Self-Attention, HGNN, and the proposed SoftHGNN under varying token sequence lengths. As illustrated in Figure \ref{fig:latency_mem}, SoftHGNN consistently exhibits significant advantages in terms of both latency and GPU memory consumption. Specifically, as the number of tokens increases, the latency and memory usage of SoftHGNN scale linearly, whereas those of Self-Attention and HGNN exhibit quadratic growth. These results indicate that SoftHGNN possesses superior computational efficiency, making it particularly suitable for modeling high-order relationships in scenarios with long sequences. This observation also empirically validates the conclusions derived from our previous computational complexity analysis.

\section{Conclusion}
% In this paper, we propose the Soft Hypergraph Neural Network (SoftHGNN), a novel plug-and-play hypergraph computation method designed to effectively capture high-order semantic relationships for general visual recognition tasks. 
\textcolor{reColor}{In this paper, we propose SoftHGNN, a lightweight plug-and-play hypergraph computation method for late-stage semantic reasoning in visual recognition.}
Unlike traditional hypergraph neural networks that rely on static and hard hyperedges, SoftHGNN introduces dynamic, feature-driven hyperedge generation with a differentiable soft participation mechanism. This innovative design significantly alleviates issues such as hyperedge redundancy and rigid semantic partitions encountered by conventional HGNNs. Furthermore, we present a sparse hyperedge selection (SeS) strategy to further enhance the representational capability of SoftHGNN while maintaining computational efficiency. Additionally, the introduced load-balancing regularizer ensures balanced activation and utilization of hyperedges throughout training, effectively addressing hyperedge selection imbalances. Extensive experiments conducted on five widely-adopted datasets across image classification, crowd counting, and object detection tasks demonstrate that the proposed SoftHGNN consistently achieves remarkable performance improvements, validating its effectiveness and generalization in capturing intricate high-order semantic relationships. We believe that SoftHGNN offers a promising foundation for advancing hypergraph computation theory and will inspire future research on more flexible, efficient, and powerful visual representation learning paradigms.

\section{Data Availability}

The datasets used in this study are available in the following public domain resources:
\begin{itemize}
    % \item CIFAR-10 and CIFAR-100: \url{https://www.cs.toronto.edu/~kriz/cifar.html}.
    \item ImageNet-100 and ImageNet-1k: \url{https://www.image-net.org/}.
    \item ShanghaiTech Part A and Part B: \url{https://github.com/desenzhou/ShanghaiTechDataset}.
    \item MS COCO: \url{https://cocodataset.org/#home}.
\end{itemize}

\section{Acknowledgment}

This work was supported by the National Natural Science Foundation of China under Grant Nos. 62501358 and U24A20252, Fundamental and Interdisciplinary Disciplines Breakthrough Plan of the Ministry of Education of China under Grant No. JYB2025XDXM504.

%%%%%%%%%%%%%%%%%%%%%%%%%%%%%%%%%%%%%%%%%%%%%%%%%%%%%%%%%%%%%%%%%%%%%%%%%%%%%%%%%%%%%%%%%%%%%%%%%%%%%%%%%%%%%%%%%%%%%%%%%%%%%%%%%%%%%%%

\bibliography{a_cite}% common bib file
%\input sn-article.bbl

\clearpage
\section{Appendix}
\subsection{Pseudo Code}
We provide the pseudo-code for SoftHGNN's soft hyperedge generation, message passing on soft hyperedges, and sparse hyperedge selection, as shown in Algorithm \ref{alg:soft_hyperedge_generation}, \ref{alg:message_passing} and \ref{alg:ses}.

% In preamble (recommended)
% \usepackage{algorithm}
% \usepackage{algpseudocode}

\begin{algorithm*}[!t]
\caption{Soft Hyperedge Generation}
\label{alg:soft_hyperedge_generation}
\begin{algorithmic}[1]
\Require Vertex/token features $X\in\mathbb{R}^{N\times D}$; learnable global prototypes $P_0\in\mathbb{R}^{M\times D}$;
context-aware offset network $\phi(\cdot)$; pre-projection matrix $W_{\mathrm{pre}}\in\mathbb{R}^{D\times D}$;
number of heads $h$ (so $D_{\mathrm{head}}=D/h$).
\Ensure Dynamic prototypes $P\in\mathbb{R}^{M\times D}$; participation matrix $A\in[0,1]^{N\times M}$.

\State \textbf{(1) Dynamic hyperedge prototype generation (Eq.(4)--(5))}
\State $f_{\mathrm{avg}} \gets \frac{1}{N}\sum_{i=1}^{N} x_i \in \mathbb{R}^{D}$
\State $f_{\mathrm{max}} \gets \max_{i=1,\dots,N} x_i \in \mathbb{R}^{D}$ \Comment{max over vertices per feature dim}
\State $f_{\mathrm{global}} \gets [f_{\mathrm{avg}}, f_{\mathrm{max}}] \in \mathbb{R}^{2D}$
\State $\Delta P \gets \phi(f_{\mathrm{global}}) \in \mathbb{R}^{M\times D}$
\State $P \gets P_0 + \Delta P \in \mathbb{R}^{M\times D}$

\State \textbf{(2) Vertex feature pre-projection + head-wise split (Eq.(6))}
\State $X_{\mathrm{proj}} \gets X W_{\mathrm{pre}} \in \mathbb{R}^{N\times D}$
\State Reshape/split $X_{\mathrm{proj}} \rightarrow X_{\mathrm{heads}}\in\mathbb{R}^{h\times N\times D_{\mathrm{head}}}$
\State Reshape/split $P \rightarrow P_{\mathrm{heads}}\in\mathbb{R}^{h\times M\times D_{\mathrm{head}}}$

\State \textbf{(3) Similarity + participation matrix (Eq.(7)--(8))}
\For{$\tau=1$ to $h$}
    \State $S^{(\tau)} \gets \dfrac{X_{\mathrm{heads}}^{(\tau)}\cdot (P_{\mathrm{heads}}^{(\tau)})^{\top}}{\sqrt{D_{\mathrm{head}}}}
    \in \mathbb{R}^{N\times M}$
\EndFor
\State $S \gets \dfrac{1}{h}\sum_{\tau=1}^{h} S^{(\tau)} \in \mathbb{R}^{N\times M}$
\For{$m=1$ to $M$}
    \State $A[:,m] \gets \mathrm{Softmax}_{\text{over vertices}}(S[:,m])$ \Comment{$A_{i,m}=\exp(S_{i,m})/\sum_{k=1}^{N}\exp(S_{k,m})$}
\EndFor
\State \Return $P, A$
\end{algorithmic}
\end{algorithm*}

\begin{algorithm*}[!t]
\caption{Message Passing on Soft Hypergraph (V$\rightarrow$E$\rightarrow$V)}
\label{alg:message_passing}
\begin{algorithmic}[1]
\Require Vertex/token features $X\in\mathbb{R}^{N\times D}$; participation matrix $A\in[0,1]^{N\times M}$;
hyperedge transform $W_e\in\mathbb{R}^{D'\times D}$; vertex transform $W_n\in\mathbb{R}^{D''\times D'}$;
activation $\sigma(\cdot)$.
\Ensure Updated vertex/token features $X'\in\mathbb{R}^{N\times D''}$.

\State \textbf{(1) Aggregation from vertices to soft hyperedges (Eq.(11)--(12), matrix Eq.(15))}
\State $F_e \gets A^{\top}X \in \mathbb{R}^{M\times D}$
\State $F'_e \gets \sigma(F_e W_e^{\top}) \in \mathbb{R}^{M\times D'}$

\State \textbf{(2) Dissemination from soft hyperedges to vertices (Eq.(13)--(14), matrix Eq.(16))}
\State $X_e \gets A F'_e \in \mathbb{R}^{N\times D'}$
\State $X' \gets \sigma(X_e W_n^{\top}) \in \mathbb{R}^{N\times D''}$

\State \Return $X'$
\end{algorithmic}
\end{algorithm*}

% In preamble (choose one style)
% \usepackage{algorithm}
% \usepackage{algpseudocode}

\begin{algorithm*}[!t]
\caption{Sparse Hyperedge Selection (SeS) and Load-Balancing}
\label{alg:ses}
\begin{algorithmic}[1]
\Require Unnormalized assignment matrix $S \in \mathbb{R}^{N \times M}$;
fixed/dynamic split sizes $M_{\text{fixed}}, M_{\text{dyn}}$ ($M=M_{\text{fixed}}+M_{\text{dyn}}$);
number of selected dynamic hyperedges $k$.
\Ensure Final participation matrix $A_{\text{total}}$ used for message passing; load-balancing loss $\mathcal{L}_{\text{LB}}$ (training only).

\State Split $S = [S_{\text{fixed}}, S_{\text{dyn}}]$, where $S_{\text{dyn}}\in\mathbb{R}^{N\times M_{\text{dyn}}}$
\For{$j = 1$ to $M_{\text{dyn}}$}
    \State Compute activation score $g_j \gets \sum_{i=1}^{N} S_{\text{dyn}}[i,j]$  \Comment{Eq.(19)}
\EndFor
\State $\text{idx} \gets \mathrm{TopK}(g, k)$  \Comment{indices of the $k$ largest scores; thresholding by the $k$-th largest is equivalent}
\State $S' \gets \mathrm{Concat}\big(S_{\text{fixed}}, \, S_{\text{dyn}}[:, \text{idx}]\big)$ \Comment{$S'\in\mathbb{R}^{N\times(M_{\text{fixed}}+k)}$}
\For{$j = 1$ to $(M_{\text{fixed}}+k)$}
    \State $A_{\text{total}}[:,j] \gets \mathrm{Softmax}_{\text{over vertices}}\big(S'[:,j]\big)$ \Comment{Eq.(20)}
\EndFor

\If{training}
    \State Construct selection mask $\mathcal{M}^{(t)} \in \{0,1\}^{M_{\text{dyn}}}$ with $\mathcal{M}^{(t)}_j = 1$ if $j \in \text{idx}$ else $0$  \Comment{Eq.(21)}
    \State Update activation probability estimate $p_j$ using recent $T$ passes  \Comment{Eq.(22)}
    \State $p_{\text{target}} \gets k / M_{\text{dyn}}$
    \State $\mathcal{L}_{\text{LB}} \gets \frac{1}{M_{\text{dyn}}}\sum_{j=1}^{M_{\text{dyn}}} (p_j - p_{\text{target}})^2$ \Comment{Eq.(23)}
\EndIf
\end{algorithmic}
\end{algorithm*}

\end{document}